\documentclass[10pt,journal,compsoc]{IEEEtran}
\usepackage{graphicx}
\usepackage{epsfig}
\usepackage{subfigure}
\usepackage{textcomp}
\usepackage{float}
\usepackage{CJK,CJKnumb,CJKulem}
\usepackage{rotating}
\usepackage{mdwlist}
\usepackage{amssymb}
\usepackage{amsfonts}
\usepackage{url}
\usepackage{algorithm}
\usepackage{algorithmic}
\usepackage{multirow}
\usepackage{amsmath}
\usepackage{xcolor}
\usepackage{lineno,hyperref}
\usepackage{epstopdf}
\usepackage{bm}
\usepackage{array}
\usepackage{wrapfig}
\usepackage{caption}
\usepackage{booktabs}
\usepackage{setspace}

\ifCLASSOPTIONcompsoc
  \usepackage[nocompress]{cite}
\else
  \usepackage{cite}
\fi
\ifCLASSINFOpdf
\else
\fi
\begin{document}

\title{Explainable Enterprise Credit Rating via\\Deep Feature Crossing Network}

\author{Weiyu Guo,
        Zhijiang Yang,
        Shu Wu,
        Fu Chen
\IEEEcompsocitemizethanks{\IEEEcompsocthanksitem Weiyu Guo, Fu Chen and Zhijiang Yang are with the Department of Information School, Central University of Finance and Economics, BeiJing 102206, P.R.China.\protect\\
E-mail: \{weiyu.guo,fu.chen\}@cufe.edu.cn, ericyang7816@foxmail.com

\IEEEcompsocthanksitem Shu Wu is with the Center for Research on
Intelligent Perception and Computing (CRIPAC), National Laboratory of
Pattern Recognition (NLPR), Institute of Automation, Chinese Academy
of Sciences (CASIA), Beijing 100864, China, and the University of Chinese
Academy of Sciences (UCAS), Beijing 100190, P.R.China.\protect\\
E-mail: shu.wu@nlpr.ia.ac.cn
}
}

\IEEEtitleabstractindextext{%
\begin{abstract}
Due to the powerful learning ability on high-rank and non-linear features, deep neural networks (DNNs) are being applied to data mining and machine learning in various fields, and exhibit higher discrimination performance than conventional methods. However, the applications based on DNNs are rare in enterprise credit rating tasks because most of DNNs employ the ``end-to-end'' learning paradigm, which outputs the high-rank representations of objects and predictive results without any explanations. Thus, users in the financial industry cannot understand how these high-rank representations are generated, what do they mean and what relations exist with the raw inputs. Then users cannot determine whether the predictions provided by DNNs are reliable, and not trust the predictions providing by such ''black box'' models. Therefore, in this paper, we propose a novel network to explicitly model the enterprise credit rating problem using DNNs and attention mechanisms. The proposed model realizes explainable enterprise credit ratings. Experimental results obtained on real-world enterprise datasets verify that the proposed approach achieves higher performance than conventional methods, and provides insights into individual rating results and the reliability of model training.
\end{abstract}

\begin{IEEEkeywords}
Feature Crossing, Explainable Learning, Credit Rating, Attention Mechanism.
\end{IEEEkeywords}}

\maketitle

\IEEEdisplaynontitleabstractindextext

%
\IEEEpeerreviewmaketitle

\IEEEraisesectionheading{\section{Introduction}\label{sec:introduction}}
The enterprise credit rating task attempts to predict the credit rating of a company by mining related data, which is critical to many financial applications, e.g., loan\cite{SME}, credit guarantee and venture investment\cite{investment}. In practice, this problem is very challenging because: (1) the information of a company is typically multi-source and heterogeneous, which results in sparse, multi-type, and high-dimensional features, and (2) accurate predictions depend on effective high-rank features because such features can add nonlinearity to the data and improve the performance of learning methods. For example, the second-rank feature, ``profit$\otimes$revenue'' often indicates the repaying ability of a company and makes sense for the enterprise credit rating task. However, it is time-consuming to obtain hand-craft high-rank features and impossible to enumerate test various combinations in polynomial fitting time, due to the input features with the property of sparse and high-dimensional. Finally, automatically generated effective high-rank features are highly efficient, which is an appealing characteristic in many real-world applications, e.g., medical treatment and fraud detection.

With the rapid development of deep learning, deep neural networks (DNNs) are receiving increasing attention in many fields, and even have become ubiquitous in a variety of applications, e.g., image processing\cite{he2016deep,Alexnet} and natural language processing\cite{bert,selfattention}, because they can extract valuable high-rank features automatically from original data without artificial feature engineering. Naturally, several previous studies\cite{addo2018credit,Bankrupt,wisa2020,golbayani2020application} have applied DNNs to the enterprise credit rating task to automatically learn the low dimensional representations of company credit. However, DNN-based forecasting models typically lack interpretability. Users neither understand the meaning of the high-rank representations generated by DNNs nor catch on the inference process of DNN models. This kind of information asymmetry and opacity shakes the norm of fair lending, which is a fundamental principle of the financial industry. In other words, users in finance field do not trust the predictions provided by a ``black box'' model. Thus, one significant challenge of using DNN models to predict enterprise credit ratings is that the ``reason codes'' should be provided to users. For example, users require an easy understood explanation of why they were denied credit, especially, when the basis for denial is the output from an opaque machine learning algorithm.

Recently, several studies\cite{AutoInt,DeepCross,xDeepFM} have been devoted to model high-rank feature interactions using DNNs to improve the performance and interpretability of forecasting models. Specifically, multiple fully-connected layers are typically used to learn the high-rank feature interactions in an implicit manner, as well as low-dimensional representations of samples. However, such kinds of methods suffer from two limitations. First, fully-connected neural networks are inefficient in terms of learning multiplicative feature interactions\cite{LatentCross}. Second, these models learn the feature interactions in an implicit manner, thus they lack good explanation to answer which feature combinations are meaningful. These limitations raise us to seek a new approach that is able to learn high-rank feature combinations explicitly for the task of enterprise credit rating task while offering a channel to penetrate ``black box'' models.

In this paper, to model the credit ratings of enterprises using deep neural network with output readable explanations, we propose a novel attention mechanism based deep neural network called DeepCross to explicitly learn effective high-rank feature combinations and predict enterprise credit ratings. In our model, to cope with sparse, multi-type, and high-dimensional features, both the categorical and numerical features are first embedded into an identical low-dimensional space, which reduces the dimension of the sparse features and allows different types of features to interact with each other. Then, inspired by self-attention, in considering a raw feature as a field query, the last high-rank representation as the key and value, we afterwards construct a series of novel feature crossing modules to explicitly learn effective patterns of high-rank feature combinations explicitly from the given dataset. As a result, we generate static explanations to penetrate the process of the proposed model's training to investigate the reliability of the model. Finally, by leveraging dual attentions, i.e., attributive and temporal attention, the proposed model adaptively indicates informative features and important time points for samples. Thus, we can provide personalized explanations for a given sample and rating pair.

Our primary contributions are summarized as follows:
\begin{itemize}
	\item We propose to study the problem of explicitly and automatically learning high-rank feature crossing in enterprise credit rating and meanwhile construct end-to-end DNN based model called DeepCross with good explainability for the target problem.  
	\item A feature crossing approach based on attentive neural network is proposed. It can  learn high-rank feature interactions automatically from both categorical and numerical input features, and support the generation of static explanations to investigate the proposed model's training process. 
	\item Dual attention modules, i.e., an attribute and temporal attention modules, are proposed to recognize the informative features and important time points of the given samples. As a result, personalized reasons of an enterprise credit rating can be furthered to insight.
	\item A series of experiments are conducted to validate the proposed model. The results demonstrate that the proposed model can obtain more precise enterprise credit ratings than conventional approaches and can provide multiple pipelines to insight the predictive process and results for users.
\end{itemize}

\section{Related Work}
The goal of this study is to propose a deep feature crossing model to obtain accurate and explainable enterprise credit ratings, thus it is relevant to three lines of study: (1) credit rating approaches for enterprises, (2) feature interactions learning techniques, and (3) attention mechanisms in the deep learning context.
\subsection{Enterprise Credit Rating}
Enterprise credit rating is an intermediary service in the financial field, which has existed for over 100 years. A mount of approaches has been proposed to handle this problem, and such approaches can be categorized into factor analysis-based methods\cite{mccrae1992an}, statistic-based methods\cite{AltmanPredicting}, and model-based methods\cite{bolton2010logistic}. 

Typically, factor analysis-based methods\cite{mccrae1992an} are usually to score the credit-related factors of an enterprise based on expert experience, which can be applied flexibly to qualitative analysis of enterprise credit. However, such methods are highly dependent on the subjective judgment of experts and lack the ability analysis enterprise credit quantitatively. Differing from factor analysis-based methods, statistic-based methods quantify the enterprise credit rating based on the company's financial indicators. For example, the Z-Score\cite{AltmanPredicting} treats the linear weighted sum of given financial indicators as the credit score of the company. The weights in the Z-Score model are calculated using the historical data of similar companies. However, such methods lack generalizability because the weights and score thresholds are fixed by experts who based on the statistical results of the historical data and their own experience to provide.

With the development of machine learning technologies, model-based approaches, e.g., logistic regression\cite{bolton2010logistic} and decision tree\cite{xia2017a}, have been used for the enterprise credit rating problem. For example, logistic regression\cite{bolton2010logistic} is often used (rather than the Z-Score\cite{AltmanPredicting}) to handle the large-scale credit rating task. In addition, the decision tree\cite{xia2017a} is also popular for the credit rating task because it can generate interpretable decision rules. However, as the features of companies become increasingly complex, the prediction performance of these models based on shallow feature representations is getting harder to be promoted. Due to the strong ability of feature representation abilities of DNNs, recent model-based credit rating approaches have transformed from traditional linear or nonlinear models to deep models\cite{Bankrupt,MATIN2019199}. Most of these models leverage recurrent neural networks (RNN) or covolutional neural networks (CNN) to learn the feature representation of credit from raw inputs, and then use multilayer feed-forward networks to predict the credit ratings. Based on this basic paradigm, although a higher accuracy can be achieved than the traditional models that use shallow feature representations, the deep models are typically considered as ``black boxes'' that cannot provide the required explanations of predictions. Thus, users neither can understand the meaning of the feature representations generated by DNNs not can catch on the inference process of DNN models. Generally, users in the field of finance do not trust predictions obtained using ``black box'' models.

\subsection{Learning Feature Crossing}
Feature crossing is a promising way to capture the interactions among raw features, and it is widely used to enhance the performance of many predictive tasks, e.g., click-through rate\cite{DeepCross,AutoInt,kdd2020} and financial analysis\cite{Bankrupt,MATIN2019199}. The results of feature crossing can indicate the cooccurrence of features and add nonlinearity to data, which can improve the performance of learning methods significantly.

Factorization Machines (FM)\cite{FM} and its extensions\cite{myFM,fmsigir} are well-known examples of learning feature interactions, which were proposed to capture the first-rank and second-rank feature interactions and have been proved effective for many tasks. However, modeling only low-rank feature interactions limits performance improvements. Thus, some recent studies\cite{AutoInt,DeepCross,kdd2020} have modeled high-rank feature interactions using DNNs to improve the power of expression. Most of these deep models follow the paradigm of embedding and stacked DNNs. Based on this paradigm, both categorical and numerical features are represented with low-dimensional vectors, and then feed-forward networks are utilized to learn the representation of high-rank feature interactions from their feature embedding. However, these approaches learn feature interactions in implicit manners, thus they lack explainability. 

In contrast, several studies have investigated learning feature interactions in explicit manners. For example, previous studies\cite{xDeepFM,DeepCross} performed explicit feature interactions by taking the outer product of features at the bit-wise or vector-wise level. However, it is important to explain which combinations are useful, because enumerating all crossing features is both impossible and unnecessary, and pursuing this leads to excessive computational complexity and may generate irrelevant or redundant feature interactions. In addition, tree-based models\cite{AutoCross,TreeEmbedding,EmbeddingForest} have been used to conduct meaningful feature interactions. However, in such methods, the training procedure was broken into multiple stages. Moreover, they rely on a certain amount of human experience and lack versatility. Finally, previous studies\cite{AutoInt,HoAFM} combined the power of embedding-based models and attention mechanisms\cite{selfattention,EMNLP19,dualattention} to learn high-rank feature interactions and identify useful feature combinations. Differing from existing studies, we explicitly model feature interactions using a self-attention mechanism in an ``end-to-end'' manner. Moreover, the proposed approach probes the learned feature combinations via lasso-based feature selection\cite{lasso}. As a result, we can learn compact and explainable feature combination patterns automatically.

\subsection{Attention Networks}
Attention was first proposed in the context of neural machine translation\cite{translate} and has been proved effective in a variety of tasks, e.g., question answering\cite{memoryNetworks}, text summarization\cite{summarization}, and recommender systems\cite{dualattention}. Recently, the self-attention mechanisms\cite{selfattention} have been used frequently to construct transformer models, from natural language processing\cite{bert} to computer vision\cite{transformorImage,transformorVideo}, which leverage multi-heads self attention\cite{selfattention} to well capture the relationships within the features. Unlike previous methods that use attention techniques to improve model accuracy, the proposed model employs the latest deep learning-based attention techniques\cite{selfattention,dualattention} to alleviate the lack of interpretability of deep learning methods. We employ attention techniques to explicitly take feature crossing and adaptively select features.

\section{Deep Feature Crossing Network}
In this section, we first give an overview of DeepCross, the proposed deep feature crossing network, which can automatically learn high-rank feature crossing for the enterprise credit rating task, and identify the meaningful feature combinations and time points. We then present a comprehensive description of how low-dimensional representations are learned explicitly for high-rank combination features without manual feature engineering, which are then used to generate accurate enterprise credit ratings.
\begin{figure*}\small
	\centering
	\includegraphics[height=0.36\linewidth,width=0.98\textwidth]{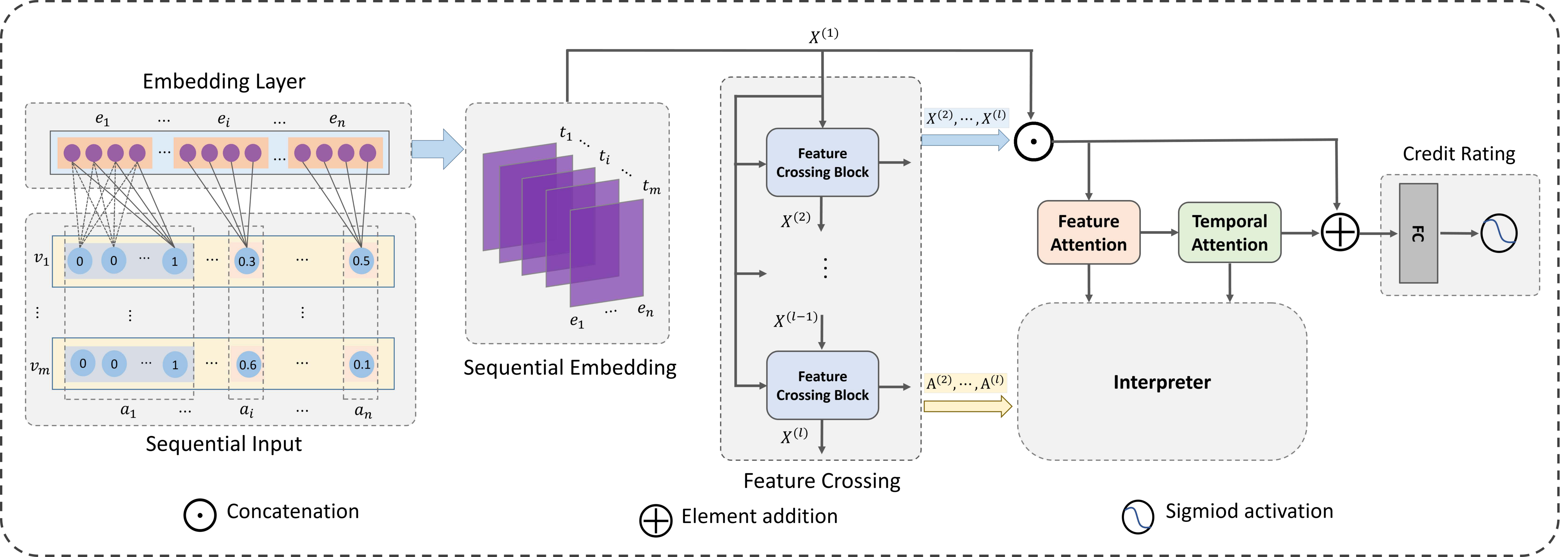}
	\caption{\small Overview of proposed model. The embedding layer projects both numerical and categorical features into the same low-dimensional feature space. The details of feature crossing block are illustrated in Fig.~\ref{fig2}, which automatically and explicitly learn the meaningful feature combinations for the task of enterprise credit rating. The feature attention module and temporal attention module are used to model the correlations of feature combinations and their temporal dependence, respectively.\vspace{-1mm}} \label{fig1}
\end{figure*}
\subsection{Overview}
The goal of the proposed approach is to map the original sparse and high-dimensional feature vector into low-dimensional spaces and model the high-rank feature interactions explicitly. As shown in Fig.~\ref{fig1}, the proposed method takes the sparse sequential data as input, followed by an embedding layer that projects both numerical and categorical features into the same low-dimensional feature space. Then, we feed the embedding of all features into a series of stacked feature crossing blocks (Fig.~\ref{fig2}), which are implemented using a self-attentive neural network and a principal component analysis (PCA) layer. For each feature crossing block, high-rank features are automatically combined using a self attentive mechanism, and the useful feature combinations are selected using a PCA layer to avoid irrelevant and redundant feature interactions, and reduce computational complexity. By stacking multiple feature crossing blocks, different ranks of feature combinations can be generated, which are the low-dimensional representations.

The outputs of the feature crossing blocks are then used to estimate the credit rating of a company. To obtain more accurate and explainable estimations, a feature attention module and a temporal attention module are stacked after the feature crossing stage. The feature and temporal attention modules are used to model the correlations of feature combinations and their temporal dependence, respectively.

\subsection{Input and Embedding Layer}
We treat the raw attributes of a company as a sequential data, and represent the $t$-th element in the sequence as a sparse vector $v_t = [a_1; a_2; ...; a_n]$, which is the concatenation of all feature fields including both numerical and categorical attributes. Here, $n$ is the total number of feature fields, $a_i$ is the attribute representation of the $i$-th feature field, $a_i$ is a one-hot vector if the $i$-th feature field is categorical (e.g., $a_1$ in Fig.~\ref{fig1}), otherwise $a_i$ is a scalar value, and the $i$-th feature field is numerical (e.g., $a_n$ in Fig.~\ref{fig1}).

The feature representations of the categorical features may be sparse and high-dimensional. A common method is to project them into a low-dimensional space. Inspired by word embeddings\cite{wordEm1,wordEm2}, we first treat each field of the categorical attributes as a lexicon and each possible value in this field as a word. Then, we can learn the low-dimensional vector representation for each categorical attribute with an embedding layer. Specifically, we represent a given categorical feature $a_j$ with a low-dimensional vector:
\begin{equation}
e_j=a_j \cdot L_j
\end{equation}
where $L_j \in \mathbb{R}^{c_{j}\times d} $ is an embedding matrix for field $j$, and $a_j$ is an one-hot vector. $c_{j}$ represents the number of categories in the field $j$, and $d$ is the dimensions of the embedding vector. In addition, considering that some categorical features can be multi-valued, e.g., the business scope of a company may cover several lines and different business lines have different contributions to the company. Therefore, if required, we process the $a_j$ to be a probability distribution vector, and the value of each dimension denotes the importance of the related category.

To realize the feature crossing can be realized between categorical and numerical features, we also represent the numerical features in a low-dimensional feature space, which is same as the feature space of categorical features. Specifically, we initialize a learnable matrix $B \in \mathbb{R}^{k\times d}$ as a basis matrix, in which the $i$-th row represents the basis of the $i$-th numerical feature in a $d$-dimensional feature space. Then, a given numerical feature $a_i$ can be expressed as follows:
\begin{equation}
e_i=a_i \cdot b_i
\end{equation}
where $b_i \in B$ is the basis vector of the numerical feature $a_i$.

By doing with the embedding layer, the sequential data are transformed as the sequencial embeddings, i.e., multiple 2D feature maps (Fig.~\ref{fig1}).

\subsection{Feature Crossing Module}\label{sec:fcm}
After both the numerical and categorical features are projected into the same low-dimensional space, we further model high-rank feature combinations in the representation space. Here, the key problem is to determine which features should be combined to form meaningful higher-rank features. Traditionally, this problem has been partly tackled by domain experts who create meaningful combinations based on their experience. In fact, human experts can only design some low-rank combinations, e.g., cost-benefit ratio, because enumerating and imagining all high-rank feature combinations is impossible to human. Thus, we tackle this problem using a neural network module inspired by self-attention mechanism and PCA.

Recently, the self-attentive network\cite{selfattention} has achieved remarkable performance in self-driven feature correlation modeling. It demonstrates superior performance when modeling arbitrary word dependency in machine translation\cite{NAACL18,EMNLP20} and the long-range dependencies of pixels in image analysis\cite{GCNet}. Here, we extend this technique to learn the correlations between different ranks of features and generate effective higher-rank feature combinations. 
\begin{figure}[!htbp]
	\centering
	\includegraphics[height=0.55\linewidth,width=0.48\textwidth]{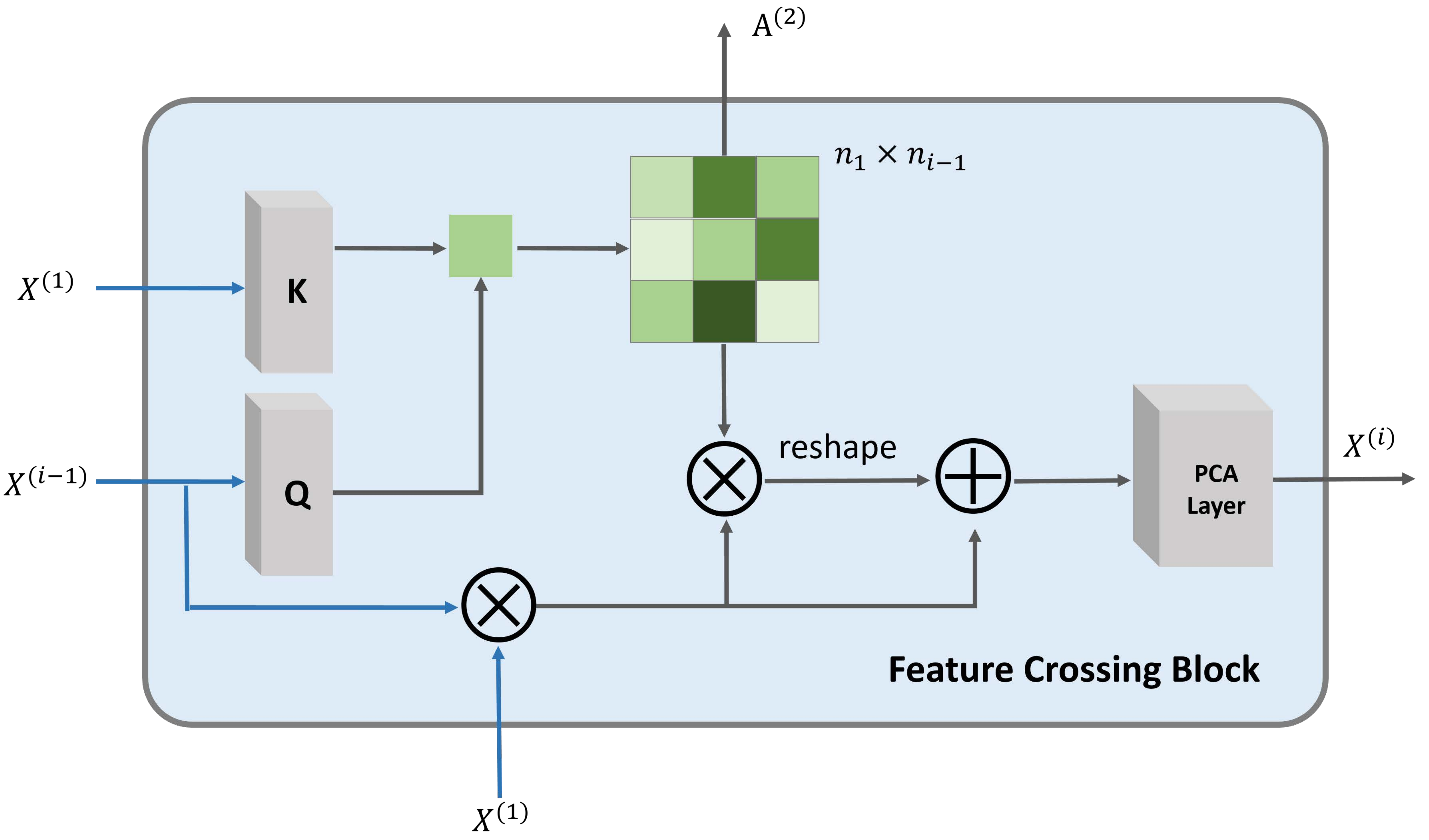}
	\vspace{-2mm}
	\caption{\small Feature crossing block, where $\bigoplus$ indicates the element wise addition and $\bigotimes$ is a Cartesian product. Grey cubes represent the convolutional networks.} \label{fig2}
\end{figure}
Specifically, we utilize the key-value attention mechanism\cite{selfattention} to dynamically determine which feature combinations are meaningful. As shown in Fig.~\ref{fig2}, taking the generation of $l$-th rank feature combinations as an example, we explain how to identify meaningful high-rank feature combinations from the candidate set, which is a set of all outer products of features. We define the first-rank features $X^{(1)}\in \mathbb{R}^{t\times n_1\times d} $ and $i-1$-th rank features $X^{(i-1)} \in \mathbb{R}^{t \times n_{i-1} \times d }$ as the input of a specific feature crossing module. We then perform a vector-level crossing product operation between $X^{(1)}$ and $X^{(i-1)}$: 
\begin{equation}
X^{(t,i)}= X^{(t,1)} \otimes  X^{(t, i-1)}
\end{equation}
where $X^{(t,i)}$ and $X^{(t,i-1)}$ are feature maps of the $t$-th time point in $X^{(i)}$ and $X^{(i-1)}$, respectively. By adopting the key-value attention mechanism, the correlations of features between $X^{(1)}$ and $X^{(i-1)}$ can be expressed as follows:
\begin{equation}
\begin{split}
& a_{m,k}=\frac{exp(\Psi(x^{(i-1)}_m,x^{(1)}_k))}{\sum_{j=1}^{n_{i-1}}exp(\Psi(x^{(i-1)}_j,x^{(1)}_k))}\\
&
\Psi(x^{(i-1)}_m,x^{(1)}_k)=\left \langle f(W_{query},x^{(i-1)}_m),  f(W_{key},x^{(1)}_k)\right \rangle
\end{split}
\end{equation}
where $\Psi(\ast,\ast)$ is an attention function that is used to define the correlation between the $m$-th feature $x^{(i-1)}_m \in X^{(t,i-1)}$ and the $k$-th feature $x^{(1)}_k \in X^{(t,1)}$. Here, we adopt the inner product of the input vectors, i.e., $\left \langle \ast,\ast \right \rangle$, as the attention function. $f(\ast,\ast)$ represents a convolution layer with a filter kernel size of $1 \times 1$. The convolution layer aggregates the temporal inputs into a feature map. $W_{query}$ and $W_{key} \in \mathbb{R}^{t}$ are the learnable parameters of the convolution layers, respectively. Finally, to learn the effectiveness of the $i$-th rank features $X^{(i)}$, we update the representation of the crossing feature $x^{(t,i)}_{m,k} \in  X^{(t,i)}$ in $d$-dimensional feature space with residual connections guided by attention coefficients $a_{m,k}$:
\begin{equation}
\widetilde{x}^{(t,i)}_{m,k}= g(a_{m,k}\cdot x^{(t,i)}_{m,k}+x^{(t,i)}_{m,k})
\end{equation}
where $g(\ast)$	is a non-linear activate function. Here, we adopt the leaky rectified linear units\cite{leaklyReLU} as the activate function, which negative slope is $0.1$.

As the feature rank increases, the number of corresponding combinations increase exponentially. This problem leads that enumerating all possible high-rank features will increase the memory and computation exponentially. In fact, only a few high-rank features are effective for target task. Therefore, we utilize a convolution network to construct PCA (Fig.~\ref{fig2}) on the candidate feature combinations. Here, a point-wise convolution neural network, which filters are conducted lasso regularization\cite{lasso} in the training stage, is used to explicitly extract the meaningful features and reduce computational costs. The implementation of the PCA layer is expressed as follows:
\begin{equation}
\arg \min_{W} L(y, f(W, X^{(i)}))+\sum_{k=1}^{c_o}  ||W_{:,k}||_{1}
\end{equation}
where $L(\ast,\ast)$ is the loss function for target label $y$. $f(\ast,\ast)$ represents a convolution neural network which the number of input channels is $c_i=n_1 \times n_{i-1}$, and the number of output channels is a hyper-parameter $c_o=n_i$. $W \in \mathbb{R}^{c_i \times c_o}$ is the learnable parameters of $1 \times 1$ convolution kernels. Due to the imposition of lasso on learnable parameter $W_{:,k}$, most of the elements in $W_{:,k}$ trend to zero after effective model training, thus we can indicate which feature combinations in $X^{(i)}$ are useful for the target task by locating the non-zero values of $W$. Finally, the representations of meaningful feature combinations $[X^{(2)},...,X^{(l)}]$ and their conditioned on attention weights $[A^{(2)},...,A^{(l)}]$ are generated using a battery of stacked $l$-$1$ feature crossing modules. Then we collect the different rank feature combinations as follows:
\begin{equation}
\widetilde{X}=X^{(1)}\odot X^{(2)}\odot \cdots \odot X^{(l)}
\end{equation}
where $\odot$ is the concatenation operator. As a result, we can obtain mixed multi-rank features about target task, and each of these features has explicit combinatorial semantics. 

\subsection{Feature Attention Module}
Once the combination features are generated in the same low-dimensional space, we further use a module of feature attention to learn the influence of different features on final enterprise credit rating. In order to adaptively calculate the attention scores of each feature, which indicate the correlations between result of enterprise credit rating and features, we assign each feature $\widetilde{X}_{:,i,:}\in \mathbf{\widetilde{X}}$ with a learnable parameter matrix $\mathbf{W}_i \in \mathbb{R}^{t \times d}$, and calculate the attention scores of features for each input sample as follows:
\begin{equation}
a_i=\frac{exp(\widetilde{X}_{:,i,:} \circledast W_i)}{\sum_{j=1}^{n}exp(\widetilde{X}_{:,j,:} \circledast W_j)}\\
\end{equation}
where $\circledast$ is an arithmetic operation that first performs element-wise multiplication, and then calculates the sum of all the results. Here, we realize this arithmetic operation using a convolution neural network, where the filter kernel size of $t \times d$, the number of its output channels is $n$ as same with the number of features. $t$ is the number of time points.

To preserve the information of previously learned combination features, we add standard residual connections to the end of this module. Formally, the output of this module is expressed as follows:
\begin{equation}
\widetilde{X}_{:,i,:}=g(a_i  \cdot \widetilde{X}_{:,i,:}+\widetilde{X}_{:,i,:})\\
\end{equation}
where $g(\ast)$	represents the rectified linear units (ReLU), which can add the nonlinearity into the proposed model.

\subsection{Temporal Attention Module}
The temporal attention module attempts to learn which time points are more informative in the sequential data, and facilitates the following credit rating in consideration of the features of critical moments. In addition, the feature fluctuations of in adjacent time points are often key patterns for credit rating forecasting, thus, we realize this attention using sliding kernels on the input sequence. Here, let $\widetilde{X}^{(t,:)}$ be the features of center time point $t$ and $s$ be the width of the sliding kernel. We calculate the attention weights for each time point in sequence as follows: 
\begin{equation}
\begin{split}
&\mathbf{X}_{s,t}=[\widetilde{X}_{t-\frac{s+1}{2},:,:},..,\widetilde{X}_{t,:,:},..,\widetilde{X}_{t+\frac{s+1}{2},:,:}]\\
&a_t=\frac{exp(\mathbf{X}_{s,t} \circledast W_s)}{\sum_{j=1}^{T}exp(\mathbf{X}_{s,j} \circledast W_s)}, t \in [1,T]\\
\end{split}
\end{equation}
where $\circledast$ is an arithmetic operation that first performs element wise multiplication, and then calculates the sum of all the results. $W_s \in \mathbb{R}^{s \times n \times d}$ is a learn-able tensor. We realize arithmetic operation $\circledast$ using 3D-convolution neural network, where convolution kernel is $W_s$ and the number of output channels is one. 


To preserve the information of previously learned combination features, we add standard residual connections to the end of this module. Formally, we obtain the following:
\begin{equation}
\widetilde{X}_{t,:,:}=g(a_t  \cdot \widetilde{X}_{t,:,:}+\widetilde{X}_{t,:,:})\\
\end{equation}
where $g(\ast)$ is adopted as the ReLUs.

\subsection{Credit Rating and Model Training}
The output of the temporal attention module is still a set of feature vectors $\{\widetilde{X}^{(i)}\}_{i=1}^{l}$, which includes all time points of the feature maps learned via the $l-1$ feature crossing modules, feature attention module, and temporal attention module. For the final credit rating prediction, we simply concatenate all feature vectors that belong to the same time point to a vector as follows:
\begin{equation}
e_t=\widetilde{X}_{t,1,:}\odot \cdots \odot \widetilde{X}_{t,n,:}
\end{equation}

Enterprise credit reflects the operation situation of the enterprise, which is influenced by both long-term and short-term operations. To balance efficiency and performance, we use GRU\cite{gru} to model the dependency of the credit ratting on long-term and short-term operations because GRU overcomes the vanishing gradients problem of RNNs and is faster than LSTM\cite{lstm}. The inputs of GRU in the proposed model are the ordered features. The formulations of GRU are expressed as follows:
\begin{equation}
\begin{split}
&u_t=\sigma (W_r\cdot[h_{t-1},e_t])\\
&z_t=\sigma (W_z\cdot [h_{t-1},e_t])\\
&\widetilde{h}_t=tanh(W_{\widetilde{h}}\cdot [u_t\ast h_{t-1},e_t])\\
&h_t=z_t\ast \widetilde{h}+ (1-z_t)\ast h_{t-1}\\
\end{split}
\end{equation}
where $\sigma(\ast)$ is the sigmoid activation function, $e_t$ is the input feature at the $t$-th time point. $W_r$, $W_z$ and $W_{\widetilde{h}}$ are the learnable parameters of GRU. $h_t$ is the $t$-th hidden states, and $\ast$ represents the element-wise product. Thus, we utilize the last hidden states $h_{T}$ of GRU as the low-dimensional feature representation of the company, and we predict the company's credit rating as follows:
\begin{equation}
\widetilde{y}=\sigma(W_{fc}\cdot h_T)
\end{equation}
where $W_{fc} \in \mathbb{R}^{k \times n}$ is the project matrix that maps the low-dimensional vector $h_{T}$ to the credit ratings $\widetilde{y}$. $n$ is the dimension of $h_{T}$, and $k$ is the number of ratings.

To train the proposed model effectively, we leverage the $L_q$ loss function\cite{gceloss} to supervise the learning process of our model. The $L_q$ loss which supervised characteristic is somewhere in between regression loss (e.g., MAE loss) and classification loss (e.g., cross entropy loss). We do this for two reasons: 1) the credit rating task can be realized by either classification or regression in practice; and 2) MAE loss typically has good generalization but less fitting ability, while cross entropy loss is the opposite. Formally, our training process is be expressed as follows:
\begin{equation}
\begin{split}
&\arg \min_{\mathbb{W}}  L_q(y_j, \widetilde{y}_j))+\sum_{i=2}^{l}\sum_{k=1}^{c_o}  ||W_{:,k}^{(i)}||_{l1}\\
&L_q(y_j, \widetilde{y_j}))=\frac{1-(y_{j} \cdot log\;\widetilde{y_j})^q}{q}
\end{split}
\end{equation}
where $\mathbb{W}$ represents the learnable parameters of the proposed model, which are updated by minimizing the total loss using gradient descent. $W_{:,k}^{(i)} \in \mathbb{W}$ is the learnable parameters of the PCA layer in the $i$-th feature crossing module (Section \ref{sec:fcm}). In addition, $\widetilde{y_j}$ is the $j$-th element in the predictive vector $\widetilde{y}$. $q \in (0,1]$ is a hyper parameter that tunes the supervised characteristic of learning between classification and regression. Note that, the loss function is equivalent to a cross entropy loss when $q \to 0$, and becomes MAE loss when $q=0$.

\section{Explainable Enterprise Rating}\label{explanation}
Through the interpreter shown in Fig.\ref{fig1}, we attempt to generate two kinds of explanations, i.e., static explanations for the rating model and individual explanations for each rating result. These explanations can rationalize the proposed model from several perspectives, i.e., training data, model reliability, and individual sample.

Owing to the imposition of lasso regularization on the parameters of PCA layers, most of the weights about feature combinations trend to zero after completing the training. Thus, given a data set and a trained model, we can identify the meaningful feature combinations for the target task as the non-zero elements in these weights. Specifically, we identify the non-zero values from the learned parameters of the PCA layers via recursive tracking. Here, assume the useful combination patterns $\Omega =\{S_{i}^{l}| l\in [1,2,...,L], i \in \Lambda^{l}\}$ have been obtained, where $\Lambda^{l}$ is the combination pattern set in which the associated parameters are non-zero in the $l$-th PCA layer. In addition, $S_{i}^{l}=\{s_j |1\leq j \leq l, s_j\in I \}$ is a feature combination pattern of the $l$-th rank, where $I$ is the set of raw features, i.e., the first-rank features. By analyzing the static explanations $\Omega$, financial experts can investigate the trained model to determine whether bias caused by the training sets is evident, and they can explore new financial indicators for the enterprise credit rating task.

To obtain comprehensible rating predictions, we further generate individual explanations for each prediction, which provides specific feature combinations and their weights by mining attention cues of the feature crossing, temporal attention and feature attention modules. Specifically, given attention score vectors $\mathbf{p}$ and $\mathbf{q}$ generated from feature attention module and temporal attention module, respectively, we generate the individual explanations by following algorithm:  
\begin{equation}
\begin{split}
&\mathbf{E}=\mathbf{r}\cdot(\mathbf{p}\otimes \mathbf{q})\\
&(e_1,e_2,..,e_k)=\textrm{topK}(\mathbf{E})
\end{split}
\end{equation}
where $\mathbf{E} \in \mathbb{R}^{t\times n} $ can be treated as the weights of the feature combinations at different time points. $\mathbf{r}$ is a distribution vector of ratings, which is treated as the probabilities on different rating classes. $(e_1,e_2,..,e_k)$ is a set which elements indicate the locations of top-k weights in matrix $\mathbf{E}$. Through the $e_i$, we can locate which feature combinations at which time points are more important to the given prediction, and we score its significance with $\mathbf{E}(e_i)$. Note that we can further identify the constituents of the given feature combination $e_i$ by parsing the useful combination patterns $\Omega$ recurrently. As a result, the interpreter can provide different explanations for different input and prediction pairs.
\section{Experiments}
In this section, we evaluate the effectiveness of the proposed approaches on real-world datasets, and attempt to answer the following questions:
\begin{itemize}
	\item How does the proposed model perform on the problem of
	enterprise credit rating problem? Is it efficient for large-scale, sparse,
	high-dimensional, and multi-type data?
	\item Are the proposed model and its outputs explainable? How to generate and understand the explanations using our proposed methods?
	\item What are the influences of different model configurations on predictive performance?
\end{itemize}

\subsection{Experimental Setup}	
\subsubsection{Datasets}
We evaluate the proposed approaches using two real-world datasets, i.e., the CH-Stocks and US-Stocks\footnote{www.kaggle.com/cnic92/200-financial-indicators-of-us-stocks-20142018} datasets.\vspace{-3mm}
\begin{figure}[!htbp]\scriptsize
	\centering
	\begin{minipage}[b]{0.52\linewidth}
		\centering
		\includegraphics[height=1\linewidth,width=1\textwidth]{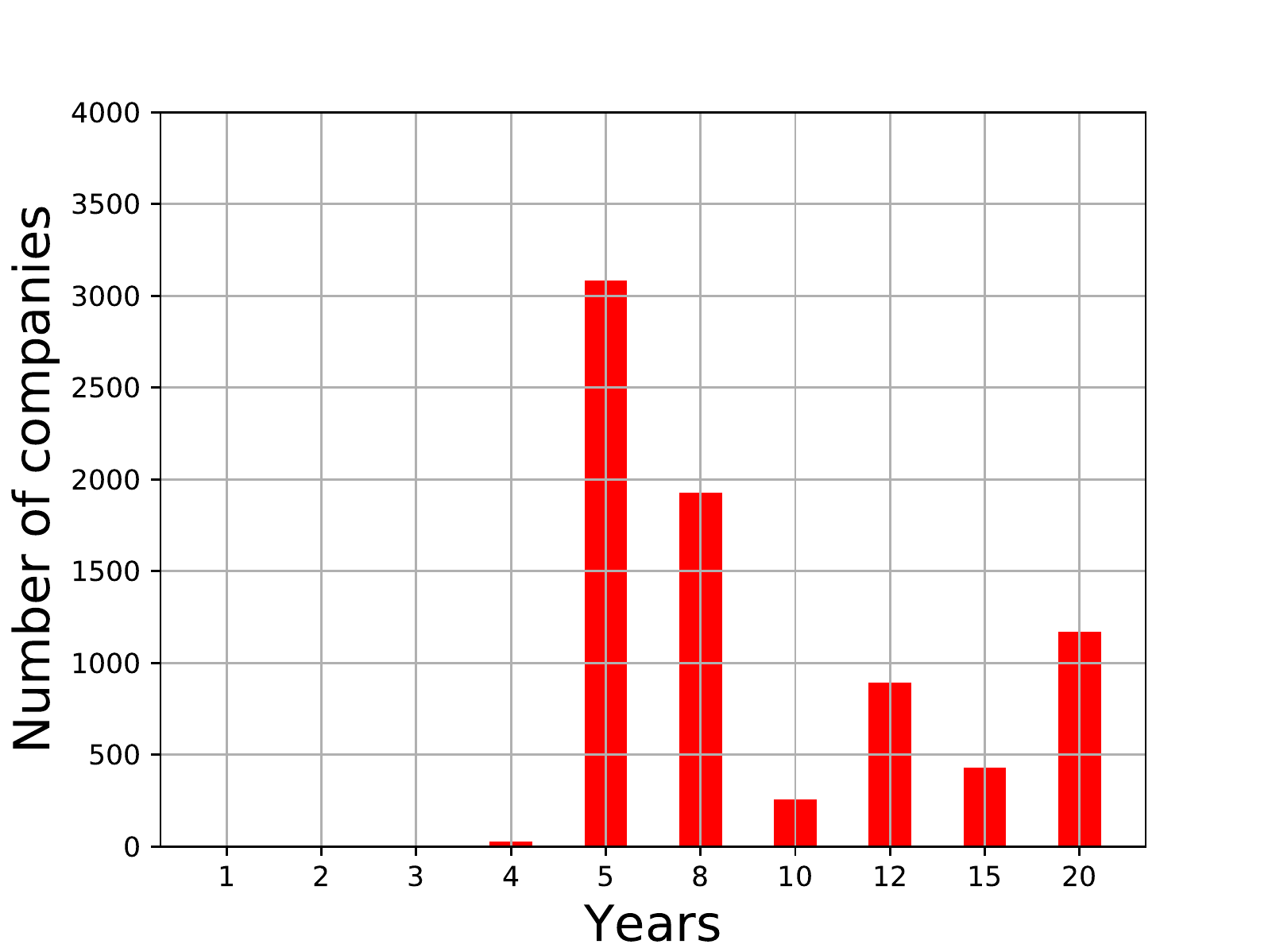}
		\vspace{-4mm}
		\caption*{(a) CH-Stocks}
	\end{minipage}%
	\begin{minipage}[b]{0.52\linewidth}
		\centering
		\includegraphics[height=1\linewidth,width=1\textwidth]{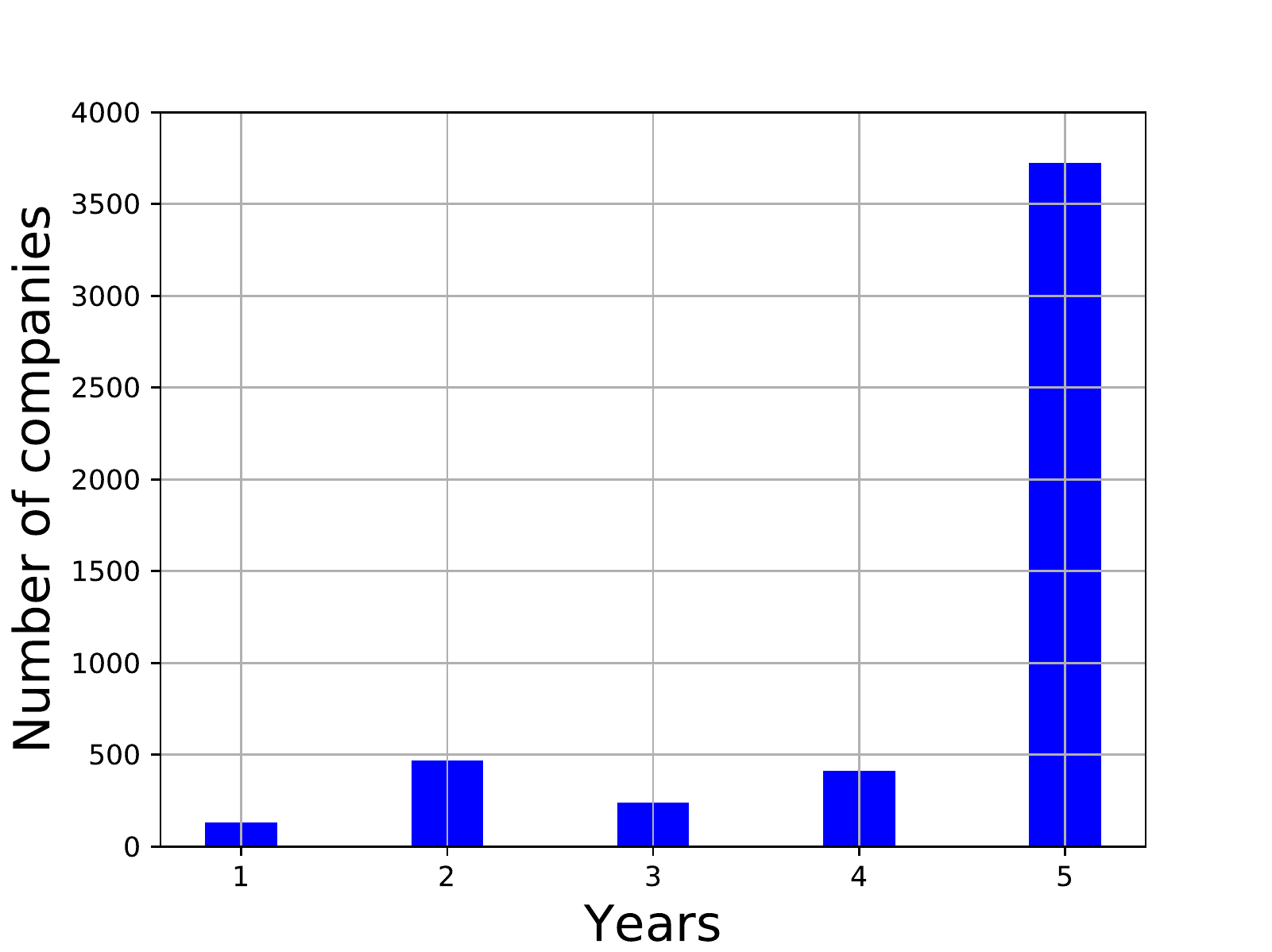}
		\vspace{-4mm}
		\caption*{(b) US-Stocks}
	\end{minipage}
	\centering
	\vspace{-4mm}
	\caption{\small Statistical distribution of the number of companies w.r.t. the time span.}
	\label{fig6}
\end{figure}

\textbf{CH-Stocks} contains the historical data of 7968 Chinese listed companies which was crawled from multiple data sources. For each company, we collected its historical data, which contains 24 first-rank financial features (Appendix \ref{appA}), from its IPO to third fiscal quarter of 2019. According to the experience of investment analysts, the revenue situation typically indicates the credit rating of a company. In this experiment, we randomly split the dataset into a training set (70\%) and a testing set (30\%), and indirectly predict the credit rating of the company by classifying whether its revenue increased in the next fiscal quarter. The testing data contained 2451 Chinese listed companies, of which 1493 of were positive samples.

\textbf{US-Stocks} contain over 200 financial indicators for all the stocks of in the US stock market yearly from 2014 to 2018. The dataset was developed to understand whether it is possible to predict the future performance of a company by looking at the financial information released in financial reports. In this experiment, we treated a company which average stock price increases in next year as a positive rating. We randomly split the dataset into a training set (70\%) and a testing (30\%) set, and indirectly predict the credit rating of the company by classifying whether its stock price increased in subsequent year.

The list of listed companies changes with time, thus, the time horizon of the data varies for each company. As shown in Fig.~\ref{fig6}, most companies in the experimental datasets have a continuous record of about 5 years. As a result, we primarily trained and tested the proposed model with five consecutive years on both CH-Stocks and US-Stocks datasets.

\subsubsection{Competing Methods}
We compared the proposed model to both deep learning-based models and conventional shallow models. Decision tree (DT), support vector machines (SVM), factorization machine (FM), gradient boosted decision trees (GBDT), Z-Score, and logistics regression(LR) are shallow models, which are often used to evaluate the credit ratings of companies. In this experiment, we concatenated all features of a company over all time periods we used as the input data of the compared shallow models (except Z-Score). AutoInt and IFR-CNN are recently proposed deep learning-based models. AutoInt was proposed for click-through rate prediction, and IFR-CNN was designed for bankruptcy prediction. However, these deep learning-based models can be transformed to perform enterprise credit rating prediction.

\textbf{Z-Score} is a classic enterprise credit rating model that treats the linear weighted sum of given financial indicators as the company's credit score, and then ratings the company by setting experience threshold scores. In this experiment, we used the Altman Z-Score model, which only considers five financial indicators with coefficients $[0.517, -0.460, 18.640, 0.388, 1.158]$, and treats the companies with Z-Score values greater than $0.9$ as the positive samples.

\textbf{LR} is a very popular model in the field of credit assessment. It is similar to the Z-Score model as it is also a linear weighted model. However, differing from the Z-Score model, the weights of LR are learned from the training data. In this experiment, the training algorithm was realized using scikit-learn\footnote{scikit-learn.org/stable} with $L_1$ penalty and the ``liblinear'' sovler. Here, we set the tolerance for the stopping criteria to $0.0001$, and kept the default values for other hyper-parameters.

\textbf{SVM}\footnote{www.csie.ntu.edu.tw/cjlin/libsvm} is a classic kernel-based model that is effective in high-dimensional spaces, even in cases where the number of dimensions is greater than the number of samples. Therefore, it is popular in the field of credit assessment field. In this experiment, the training algorithm was realized using ``libsvm'', which is ``nu-SVC'', the ``rbf'' kernel type, and the defaults configurations for other settings.

\textbf{DT} is a representative algorithm for classification that is commonly used in finance. In this experiment, the maximum tree depth was set to $5$, and the training algorithm was realized using scikit-learn. Note that the criterion for building the tree is evaluating the Gini impurity. 

\textbf{FM} is a general framework that uses factorization techniques to model second-rank feature interactions and provide high prediction accuracy. In this experiment, we used libFM\footnote{www.libfm.org} and the adaptive SGD learning method with learning rate $0.1$, iteration times $500$, and the defaults configurations for other settings to learn the second-rank feature interactions of the financial indicators and predict the credit ratings of companies.   

\textbf{GBDT} is a boosting learning technique for both regression and classification problems that produces a prediction model in the form of an ensemble of weak decision trees. In this experiment, the training algorithm was realized using scikit-learn. We set the number of estimators to $100$, and the number of random states was set $10$. Default values were used for other hyper-parameters.

\textbf{AutoInt}\footnote{github.com/shichence/AutoInt} automatically learns the high-rank feature interactions using multi-head self-attention, which can map both the numerical and categorical features into the same low-dimensional space. In this experiment, we set the number of heads and blocks to $2$ and $3$, respectively, and the block shape was set to $[64, 64, 64]$. To facilitate fair comparison, we employed AutoInt to obtain the feature representation of a sample, and then used our feature attention module and temporal attention module to generate prediction.

\textbf{IFR-CNN} transforms the bankruptcy prediction to be the task of image classification by generating matrices of financial ratios. In this experiment, we realize the matrix generation approach of IFR-CNN, and generate the matrices of financial ratios by using the data of fiscal quarter or year. Then, a binary classifier based on googlenet\cite{googlenet} realized by torchvision\footnote{github.com/pytorch/vision} is trained and tested by using the generated matrices. Its task is predicting the next situation based on current data. In other words, it only leverages data from a single time point to make predictions.

\subsection{Evaluation Criteria}
\begin{table}[!htbp]\small
	\caption{\small Confusion matrix of classifiers in this experiment. $y=1$ indicates that the model gives the company a positive outlook, $y=0$ represents negative outlook. $g=1$ indicates that the actual credit rating of the company is positive, and $g=0$ is a negative outlook.\vspace{-1mm}}\label{tab1}
	\centering
	\begin{tabular}{p{10 pt}| p{25 pt}<\centering|p{68 pt}<\centering|p{68 pt}<\centering}
		\specialrule{0.2em}{1.1pt}{1.1pt}
		\multicolumn{2}{c|}{\multirow{2}{*}{Confusion Matrix}}  &\multicolumn{2}{c}{Ground Truth}\\
		\specialrule{0em}{1.1pt}{1.1pt}
		\cline{3-4}\multicolumn{1}{c}{~} & ~ & $g=1$ & $g=0$\\
		\hline
		\specialrule{0em}{1.1pt}{1.1pt}
		\multicolumn{1}{c|}{\multirow{2}{*}{Prediction}} &	$y=1$ & $\;$ True Positive(TP) $\;$ & $\;$ False Positive(FP)$\;$ \\	
		\specialrule{0em}{1.1pt}{1.1pt}
		\cline{2-4}~&$y=0$ & $\;$ False Negative(FN)$\;$  &$\;$ True Negative(TN)$\;$ \\
		\specialrule{0.2em}{1.1pt}{1.1pt}
	\end{tabular}
\end{table}

In binary classification problems, the classifiers are likely to obtain four types of predictions, i.e., true positives (TP), false positives (FP), false negatives (FN), and true negatives (TN). Therefore, we first defined the confusion matrix for our experiments as shown in Table \ref{tab1}. To achieve a comprehensive evaluation, the predictive performance of the compared models was evaluated using several indicators, e.g., accuracy, type \uppercase\expandafter{\romannumeral1} and \uppercase\expandafter{\romannumeral2} errors, and area under the ROC curve (AUC). Based on the confusion matrix defined in Table~\ref{tab1}, the accuracy, type \uppercase\expandafter{\romannumeral1} and \uppercase\expandafter{\romannumeral2} errors can be defined as follows:
\begin{equation}
\begin{split}
&Acc=\frac{TP+TN}{TN+FP+FN+TP}\\
& Err_{1}=\frac{FP}{TN+FP}\\
& Err_{2}=\frac{FN}{FN+TP}
\end{split}
\end{equation}
where, $Acc$ indicates the accuracy of labels prediction by a given model. $Err_{1}$ and $Err_{2}$ are type \uppercase\expandafter{\romannumeral1} error and type \uppercase\expandafter{\romannumeral2} error, respectively. They can investigate the performance of binary classification with the viewpoint of different categories. 

To avoid the evaluation error caused by sample imbalance, the AUC is utilized to further evaluate the quality of models in this experiment, where ROC is a comprehensive indicator reflecting continuous variables of True Positive Rate and False Positive Rate continuous variables. The AUC value is between $0.5$ and $1$, and a higher value is better.
\begin{table*}
	\caption{\small Performance of enterprise credit rating prediction of different models. In this experiment, our model was trained to be a third-rank model both on CH-Stocks and US-Stocks, which the output dimension of the embedding layer is $64$. The output dimensions of PCA layers in our stacked feature crossing modules are $[128,64,32]$, which indicate the numbers of feature combinations that are retained in different rank feature crossing. DeepCross and AutoInt were trained and tested by using the data of $5$ consecutive time points on US-Stocks and $15$ consecutive time points on CH-Stocks.}\label{tab2}
	\centering 
	\begin{tabular}{p{45 pt}<\centering|p{80 pt}<\centering|p{30 pt}<\centering|p{30 pt}<\centering|p{30 pt}<\centering|p{30 pt}<\centering|p{30 pt}<\centering|p{30 pt}<\centering|p{30 pt}<\centering|p{30 pt}<\centering p{30 pt}}
		\specialrule{0.2em}{1.1pt}{1.1pt}
		\multirow{2}{*}{Model class}&
		\multirow{2}{*}{Models}&
		\multicolumn{4}{c|}{CH-Stocks}&	
		\multicolumn{4}{c}{US-Stocks}\\ 
		\specialrule{0em}{1.1pt}{1.1pt}
		\cline{3-10}~ & &Acc &AUC &Err$_1$ &Err$_2$&Acc &AUC &Err$_1$ &Err$_2$\\ \hline
		\specialrule{0em}{1.1pt}{1.1pt}
		\multirow{2}{*}{First-rank}&Z-Score\cite{AltmanPredicting} &0.6359 &-- &0.3033 &0.7523&0.6987 &-- &0.2915 &0.3660\\
		\specialrule{0em}{1.1pt}{1.1pt}
		~&LR\cite{logisticregression} &0.8464 &0.9252 &0.1678 &0.0883 & 0.7256 &0.7813 &0.2682 &0.2901\\
		~&SVM\cite{svm2011} &0.7926 &0.8458 &0.2223 &0.0928 &0.7269 &0.7791 &0.2879 &0.2187\\
		\hline
		\specialrule{0em}{1.1pt}{1.1pt}
		\multirow{3}{*}{High-rank}~&FM\cite{libfm} &0.8244 &-- &0.1837 &0.1373 &0.7249 &--&0.2632 &0.3029\\
		\specialrule{0em}{1.1pt}{1.1pt}
		&GBDT\cite{gdbt} &0.9526 &0.9713 &0.0397 &0.0521 &0.7503 &0.8279 &0.2546 &0.228\\
		\specialrule{0em}{1.1pt}{1.1pt}
		&DT\cite{dtrees}&0.9314 &0.9623 &0.0551 &0.0846 &0.7342 &0.8035 &0.2856  &\textbf{0.1889}\\ \hline
		\specialrule{0em}{1.1pt}{1.1pt}
		\multirow{3}{*}{Deep-rank}~ &AutoInt\cite{AutoInt} &0.9612 &0.9736 &0.02758  &0.0416  &0.7497 &0.7869 &0.2457 &0.2653\\
		\specialrule{0em}{1.1pt}{1.1pt}
		&IFR-CNN\cite{Bankrupt}  &0.6716 &0.7072 &0.3275 &0.3292 &0.7127 &0.7012 &0.3027 &0.3662\\
		\specialrule{0em}{1.1pt}{1.1pt}
		&\textbf{DeepCross} &\textbf{0.9801} &\textbf{0.9955} &\textbf{0.0166} &\textbf{0.0275} &\textbf{0.7723} &\textbf{0.8341} &\textbf{0.2303} &0.2537\\
		\specialrule{0.1em}{1.1pt}{1.1pt}
	\end{tabular}
\end{table*}
\subsection{Quantitative Analysis}	
\subsubsection{Evaluation of Effectiveness}	
We summarize the quantitative results of company credit ratings obtained by different models in Table~\ref{tab2}. The following observations can be obtained: (1) LR and SVM which are a machine learning based linear model significantly outperformed the statistics analysis-based model, i.e., Z-Score model, because they can adaptively fit the distribution of the given dataset, which may be more suitable to the company credit rating task on large-scale data. Besides, the performance of Z-Score model on US-Stocks dataset was better than CH-Stocks dataset. The reason for this phenomenon is that the weights of Z-Score model in this experiment obtained from corporate statistics in advanced economies, which may be not suite for the Chinese situation. (2) GBDT and DT, which explore high-rank feature engineering, consistently outperformed the first-rank approaches by a large margin on all datasets, which indicates that using only first-rank features may be insufficient in company credit rating prediction. (3) Benefiting from the feature engineering capabilities of DNNs and the attention mechanism, the DeepCross and AutoInt typically achieved better performance than other models. (4) The proposed model, i.e., DeepCross, obtained the best performance, which indicates that using feature crossing modules to explore deeper-rank feature interactions is crucial. Note that the proposed model shares the same structures as AutoInt (except the setup of the feature crossing modules). (5) The deep learning-based model IFR-CNN did not consistently show advantages compared to some of the shallow models. The reason for this phenomenon may be that only leveraging data from a single time point is not sufficient for the company credit rating prediction task. The company credit changes with time and it may be a sequential process.

In summary, the proposed model outperformed all compared models. Compared to the most competitive baseline i.e., AutoInt, the proposed model could explore deeper-rank feature interactions with similar resource consumption and is more efficient during online inference. This advantage is gained through the stacked feature crossing modules, which first perform explicit feature crossing via the vectorized Caresian product, and then perform PCA using a one-dimensional convolutional network.

\subsubsection{Influence of Different Rank}
\begin{figure}[!htbp]\small
	\centering
	\begin{minipage}[b]{0.5\linewidth}
		\centering
		\includegraphics[height=1\linewidth,width=1\textwidth]{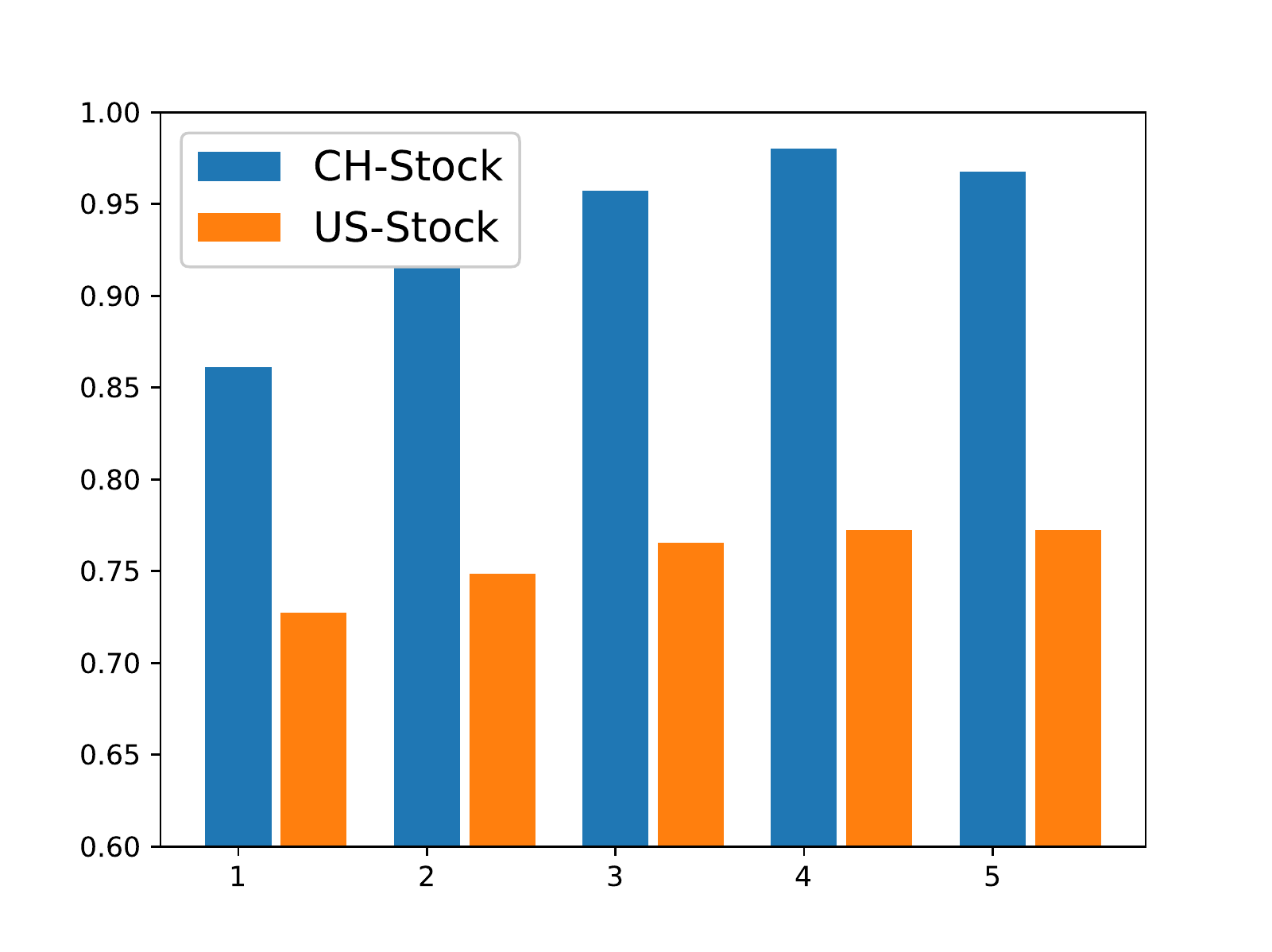}
		\vspace{-7mm}
		\caption*{\small (a) Accuracy}
	\end{minipage}%
	\begin{minipage}[b]{0.5\linewidth}
		\centering
		\includegraphics[height=1\linewidth,width=1\textwidth]{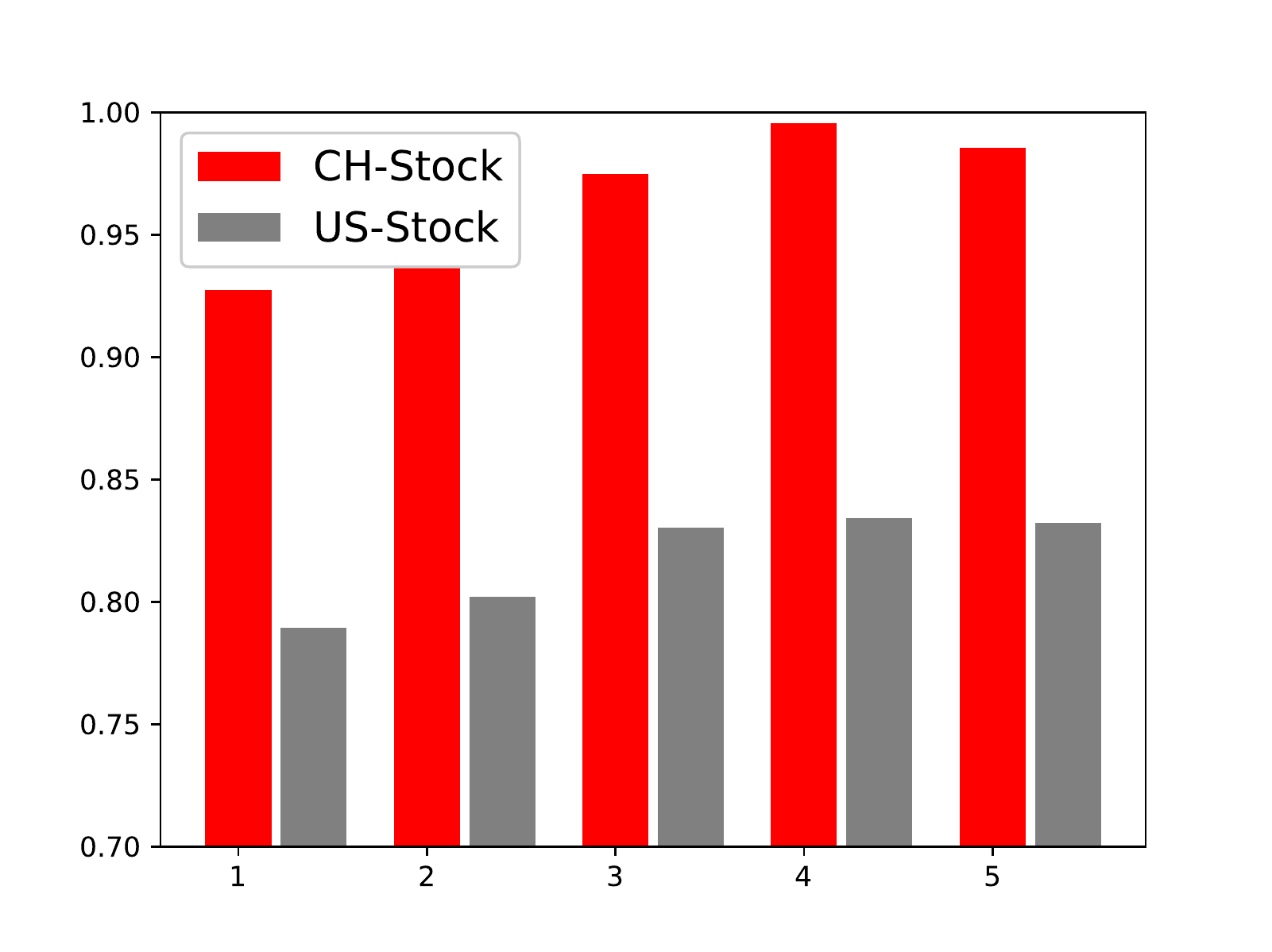}
		\vspace{-7mm}
		\caption*{\small (b) AUC}
	\end{minipage}
	\centering
	\vspace{-4mm}
	\caption{\small Performance w.r.t. the rank of the model.}
	\label{fig3}
\end{figure}

The proposed model learns high-rank feature combinations by stacking multiple feature crossing modules. We investigated the performance of the proposed model in terms of parameter $l$, which is the rank of feature combinations. As shown in Fig.~\ref{fig3}, the performance typically increased as we increased the rank of the proposed model because higher-rank feature crossing means that more feature combinations are used for prediction. However, the results obtained on two datasets differ somewhat. When the range of feature rank over $4$, the performance of the proposed model on the CH-Storcks dataset began to decrease. The reason for this reduced performance was likely by the fact that the number of first-rank features in CH-Storcks dataset is small, and the fourth-rank and above features contain too many invalid feature combinations. As a result, the number of training samples may be relatively small compared to the feature dimension, which caused the proposed model to exhibit over-fitting.	\vspace{-4mm}
\begin{figure}[!htbp]
	\centering
	\begin{minipage}[b]{0.5\linewidth}
		\centering
		\includegraphics[height=1\linewidth,width=1\textwidth]{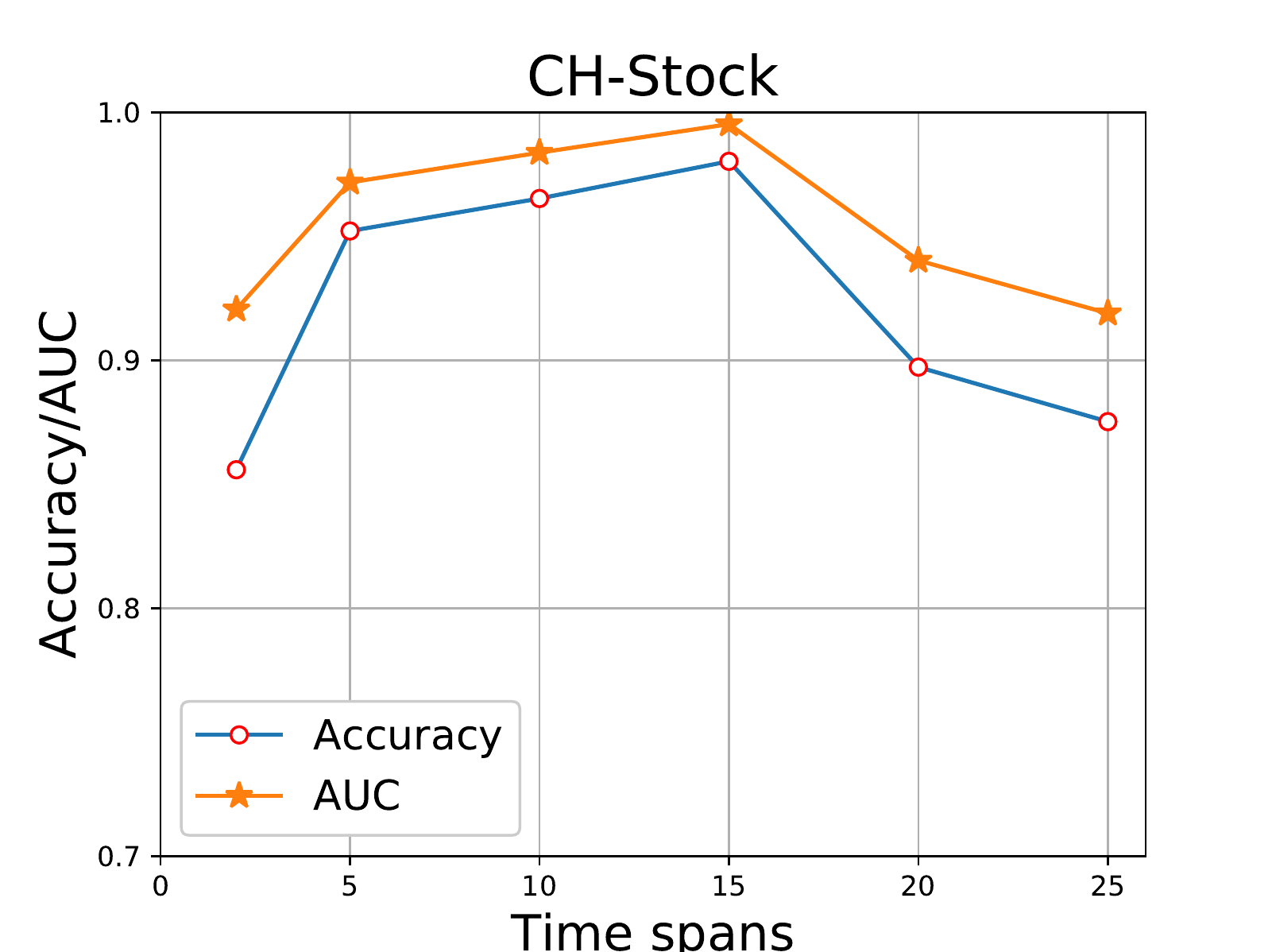}\vspace{2mm}
		\includegraphics[height=1\linewidth,width=1\textwidth]{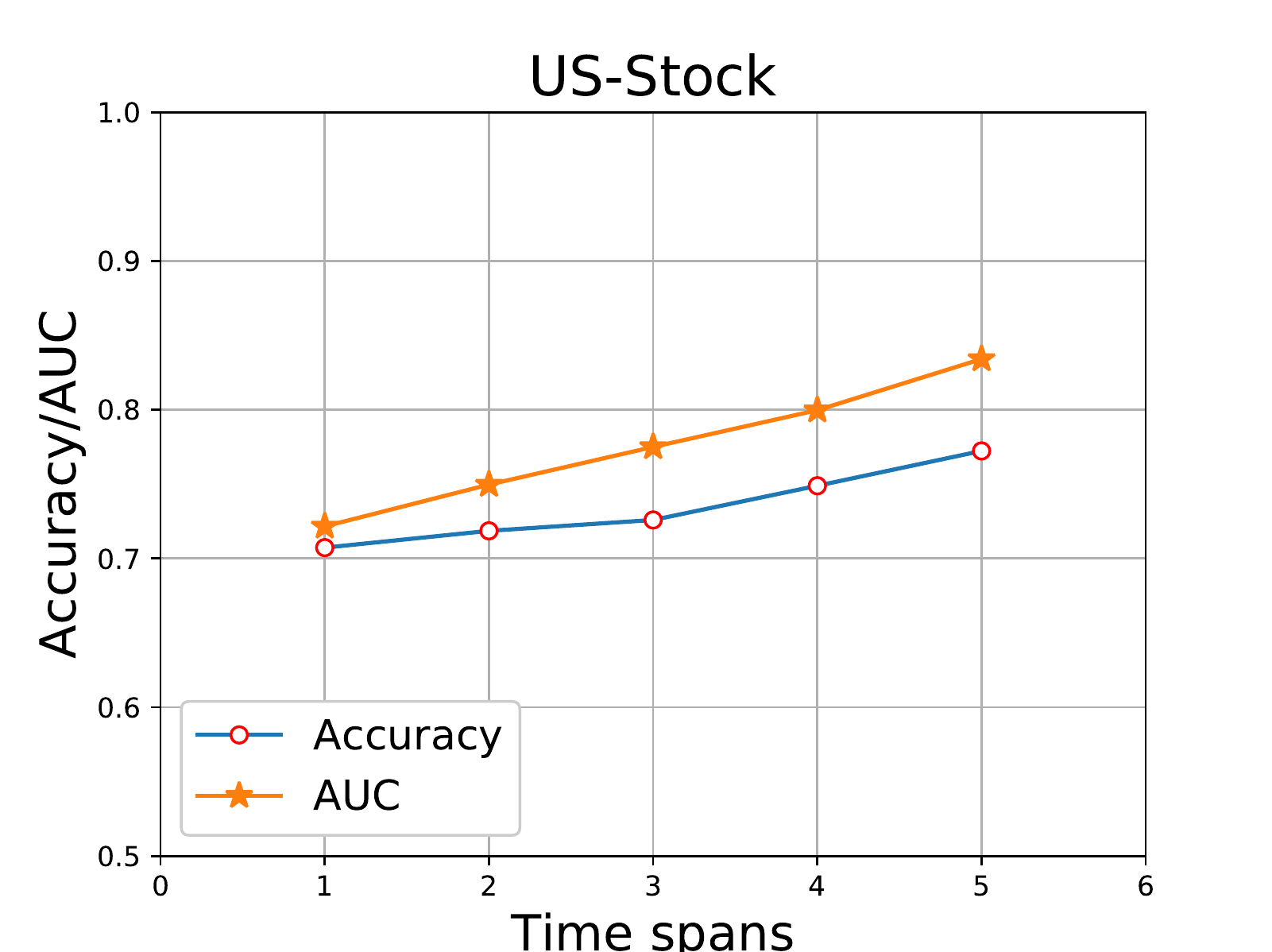}
		\vspace{-4mm}
		\caption*{\small (a) Testing phase}
	\end{minipage}%
	\begin{minipage}[b]{0.5\linewidth}
		\centering
		\includegraphics[height=1\linewidth,width=1\textwidth]{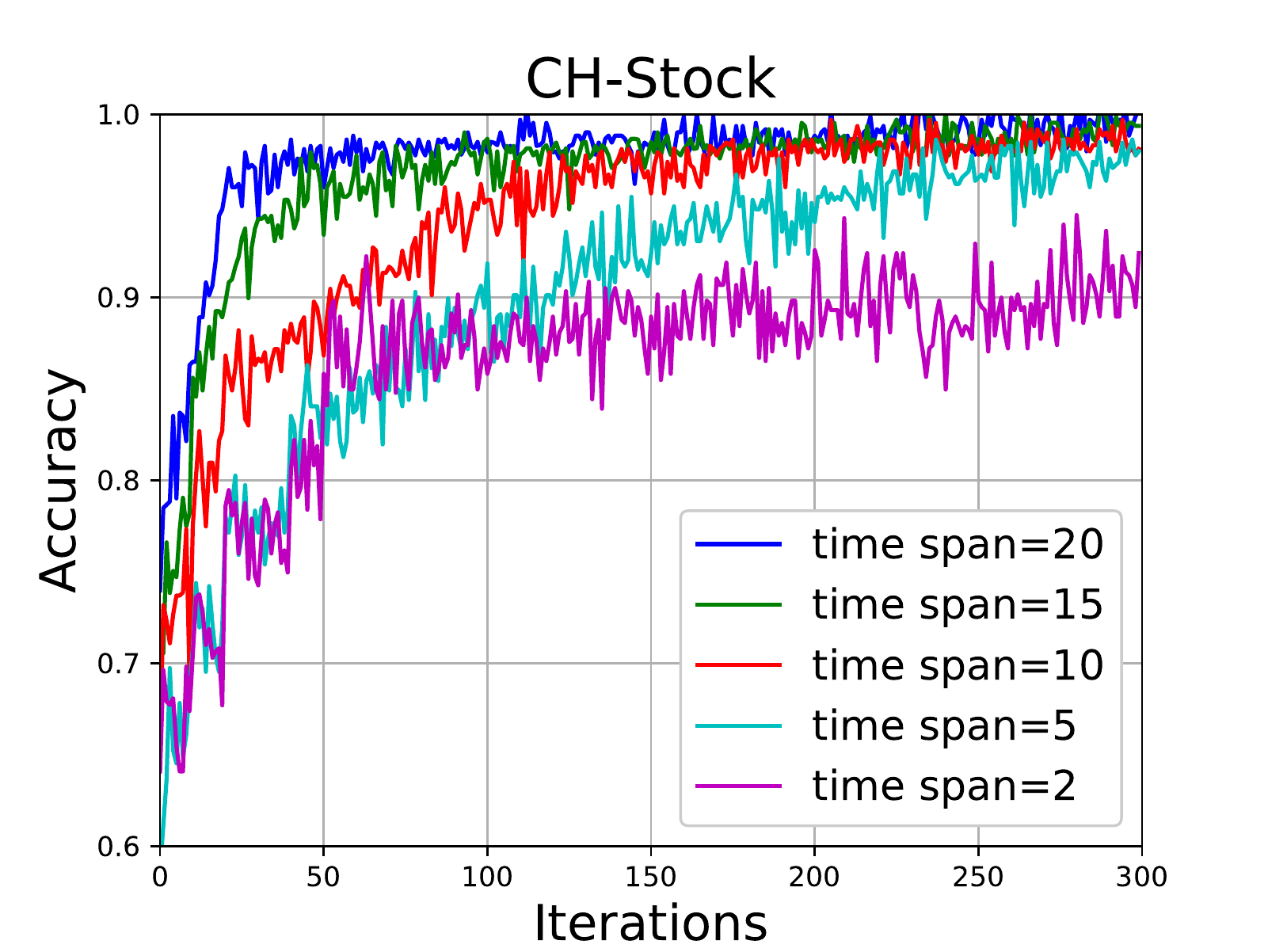}\vspace{2mm}
		\includegraphics[height=1\linewidth,width=1\textwidth]{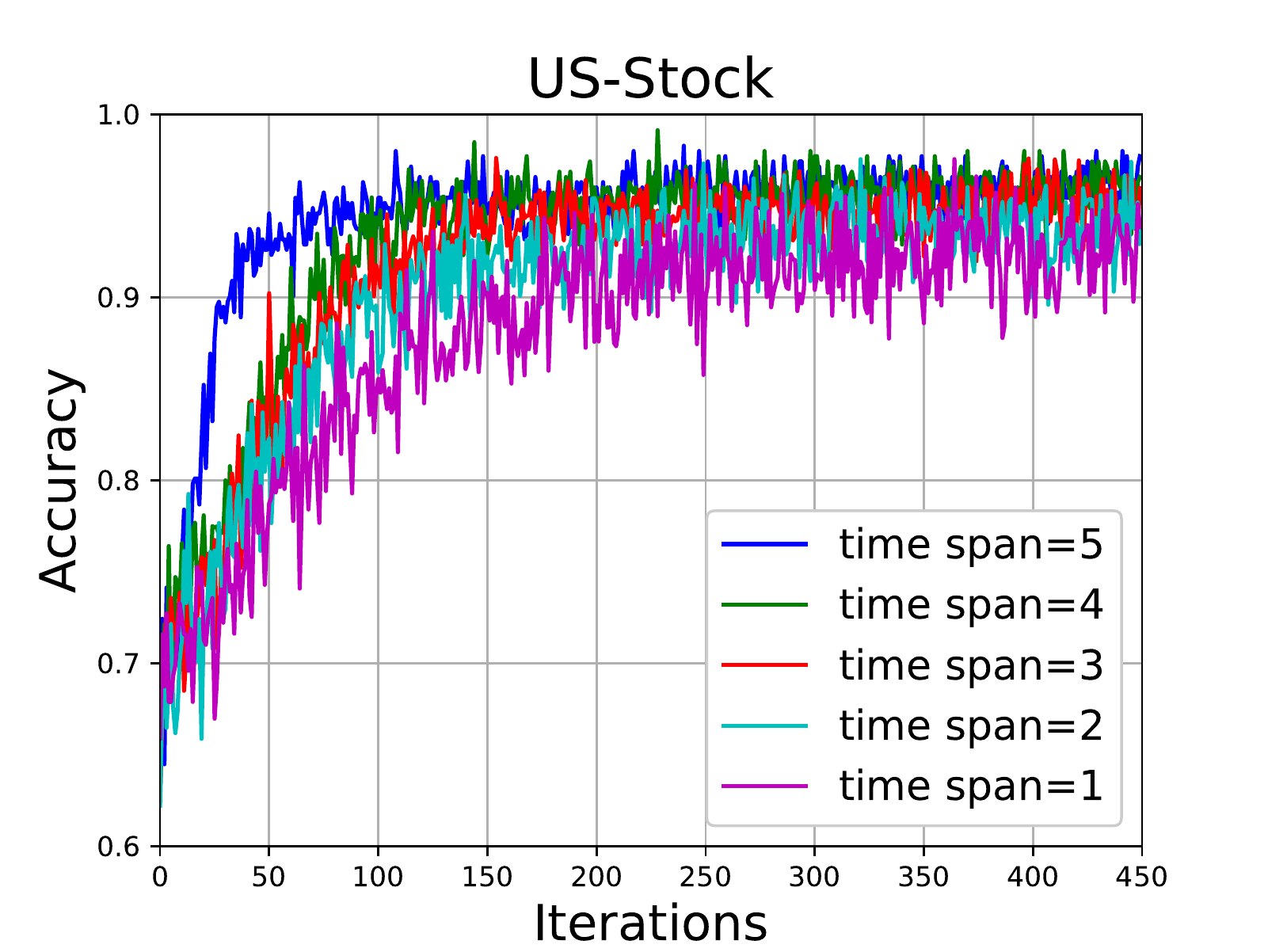}
		\vspace{-4mm}
		\caption*{\small (b) Training phase}
	\end{minipage}
	\centering
	\caption{\small Influence of different time spans in training and testing phases. In training phase, we adopted the method of stochastic gradient descent with $0.001$ learning rate to train our models.}
	\label{fig4}
\end{figure}	
\subsubsection{Influence of Different Time Spans}
We predict the credit ratings of companies by sequential modeling. Therefore, we investigate the feature of our model in training and testing phase w.r.t. the parameter $t$, which is the time spans of data used to training and testing. As shown in Fig.~\ref{fig4}, accuracy and AUC firstly increase as we increasing the time span on testing phase since credit rating of a company is temporal evolution. However, when the time span exceeds a certain range, the performance of our model begins to decrease on CH-Stocks. The reason of this phenomenon is probable that the increase of time span leads to significant reduction of training samples in CH-Stocks, which causes the trained model to suffer from over-fitting. 

In addition, different from the testing phase, with any time spans settings, the proposed model achieved convergence on the training datasets, and the convergence speed of training was accelerated as the time span was increased. Note that as the time span increases, the number of training samples in datasets decreases. This phenomenon further indicates that the number of training samples can affect the fitting performance of the model, and increasing the number of samples may be an effective way to improve the proposed model's performance.
\begin{table*}[!htbp]
	\centering
	\caption {\small Most important feature combinations on CH-Stocks dataset. The $f_i$ in table indicates a first-rank feature, which special semantics can be queried in Appendix \ref{appA}. $l_i$ is the rank of features.\vspace{-2mm}}
	\begin{tabular}{c|c|c|c|c|c|c|c|c}
		\specialrule{0.2em}{1.1pt}{1.1pt}
		\multirow{2}{*}{Top-K}&
		\multicolumn{2}{c|}{$l_1$}&
		\multicolumn{2}{c|}{$l_2$}&	
		\multicolumn{2}{c|}{$l_3$}&
		\multicolumn{2}{c}{$l_4$}\\ 
		\specialrule{0em}{1.1pt}{1.1pt}
		\cline{2-9}~&feature &weight &feature &weight&feature &weight&feature &weight\\ \hline
		\specialrule{0em}{1.1pt}{1.1pt}
		1&$f_{18}$&0.3852&$f_{11}$,$f_{18}$&0.1995&$f_{6}$,$f_{18}$,$f_{20}$&0.0177&$f_{4}$,$f_{4}$,$f_{7}$,$f_{5}$&0.0500\\
		\specialrule{0em}{1.1pt}{1.1pt}
		2&$f_{15}$&0.0878&$f_{18}$,$f_{20}$&0.1599&$f_{0}$,$f_{3}$,$f_{22}$&0.0177&$f_{0}$,$f_{12}$,$f_{18}$,$f_{21}$&0.0499\\
		\specialrule{0em}{1.1pt}{1.1pt}
		3&$f_{13}$&0.0809&$f_{6}$,$f_{18}$&0.1594&$f_{18}$,$f_{18}$,$f_{21}$&0.0177&$f_{3}$,$f_{3}$,$f_{9}$,$f_{15}$&0.0250\\
		\specialrule{0em}{1.1pt}{1.1pt}
		4&$f_{12}$&0.0667&$f_{3}$,$f_{15}$&0.1198&$f_{9}$,$f_{12}$,$f_{13}$&0.0177&$f_{11}$,$f_{15}$,$f_{17},f_{19}$&0.0250\\
		\specialrule{0em}{1.1pt}{1.1pt}
		5&$f_{10}$&0.0593&$f_{12}$,$f_{18}$&0.1198&$f_{8}$,$f_{0}$,$f_{22}$&0.0177&$f_{6}$,$f_{10}$,$f_{12}$,$f_{24}$&0.0250\\ 
		\specialrule{0.1em}{1.1pt}{1.1pt}
	\end{tabular}
	\label{tab3}
\end{table*}
\begin{figure*}[!htbp] 
	\begin{minipage}[b]{0.5\textwidth}
		\centering
		\includegraphics[height=0.65\linewidth,width=0.9\textwidth]{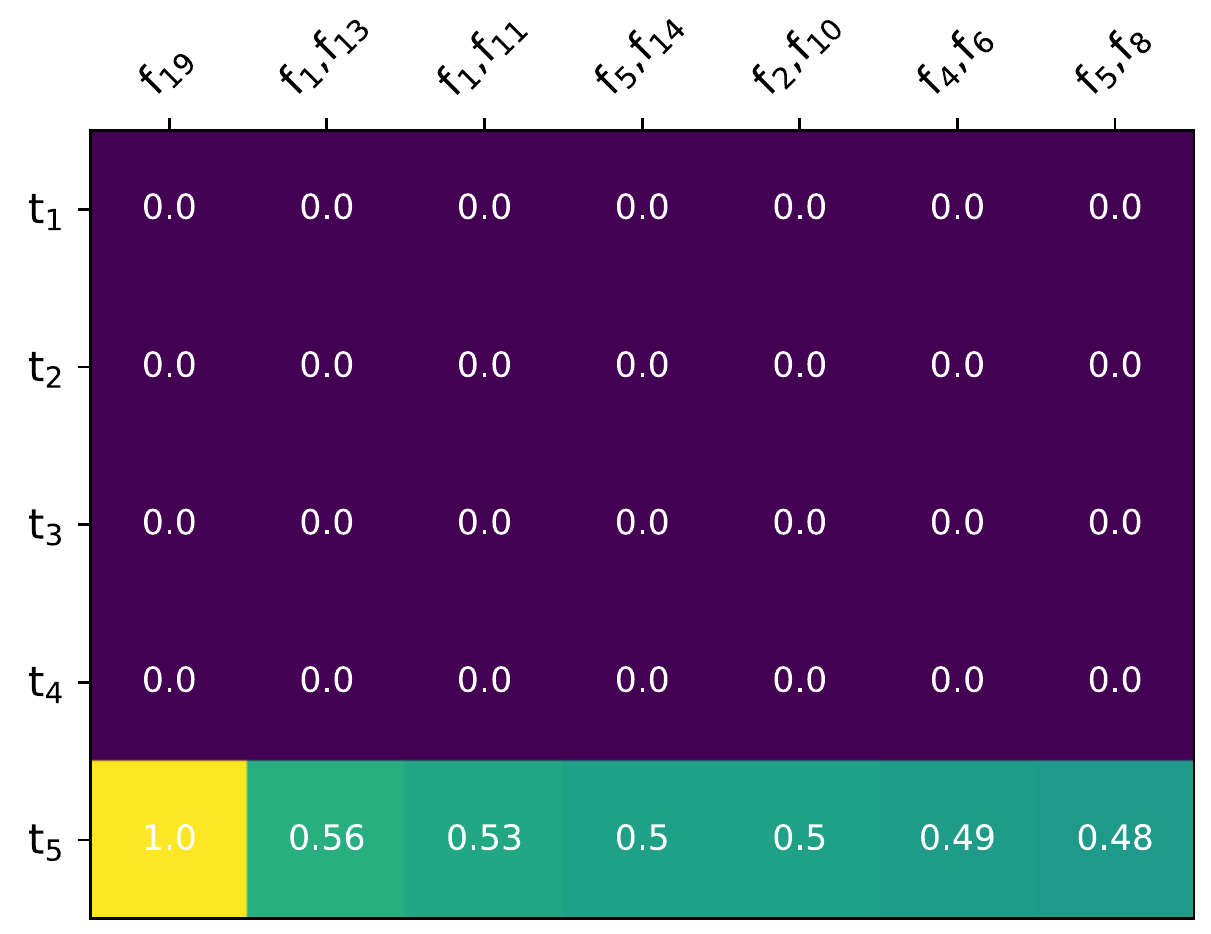}\vspace{1em}
		\includegraphics[height=0.7\linewidth,width=0.9\textwidth]{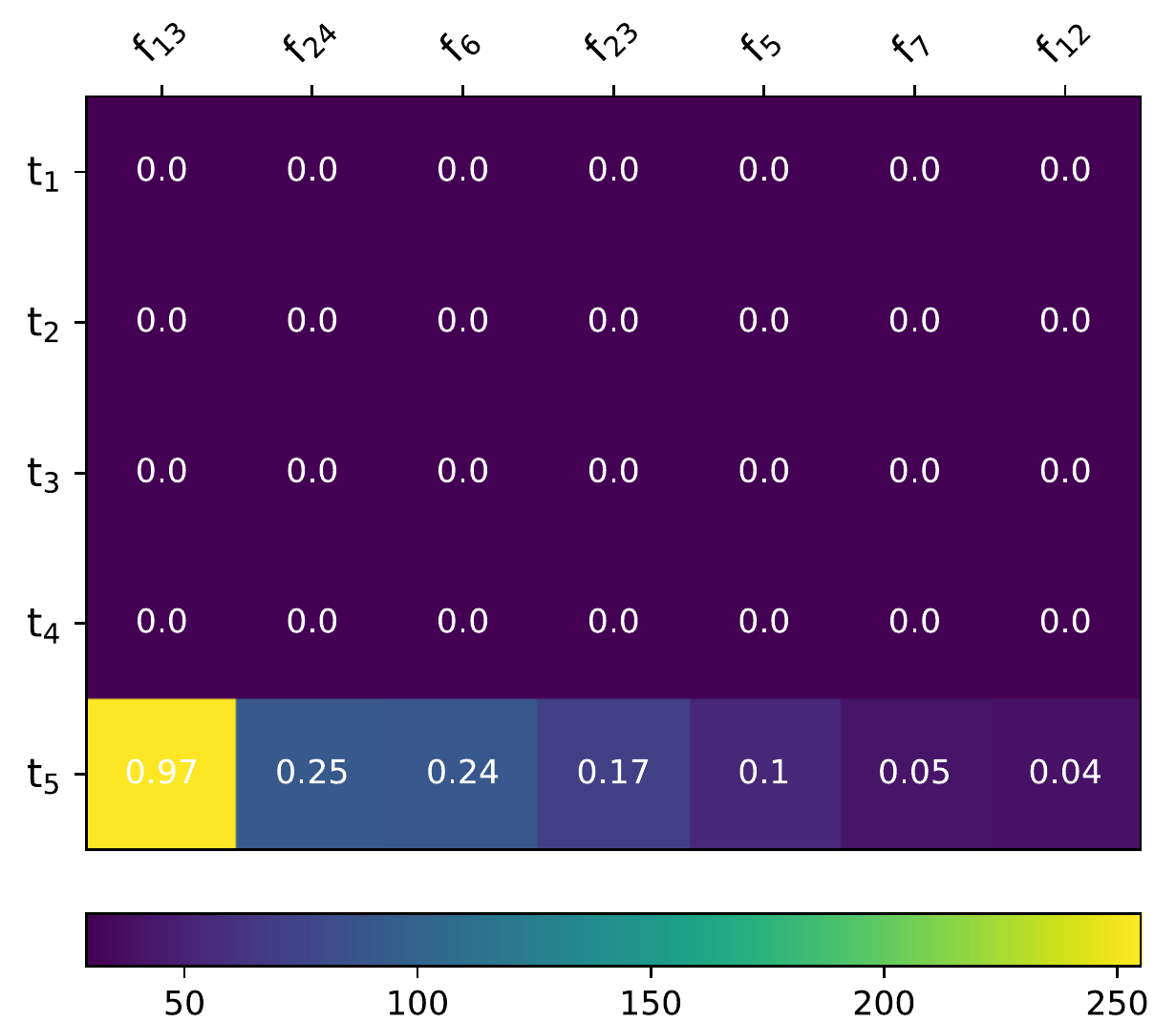}
	
		\caption*{\small (a) Positive instances.\vspace{-2mm}}
	\end{minipage}
	\begin{minipage}[b]{0.5\textwidth}
		\centering
		\includegraphics[height=0.65\linewidth,width=0.9\textwidth]{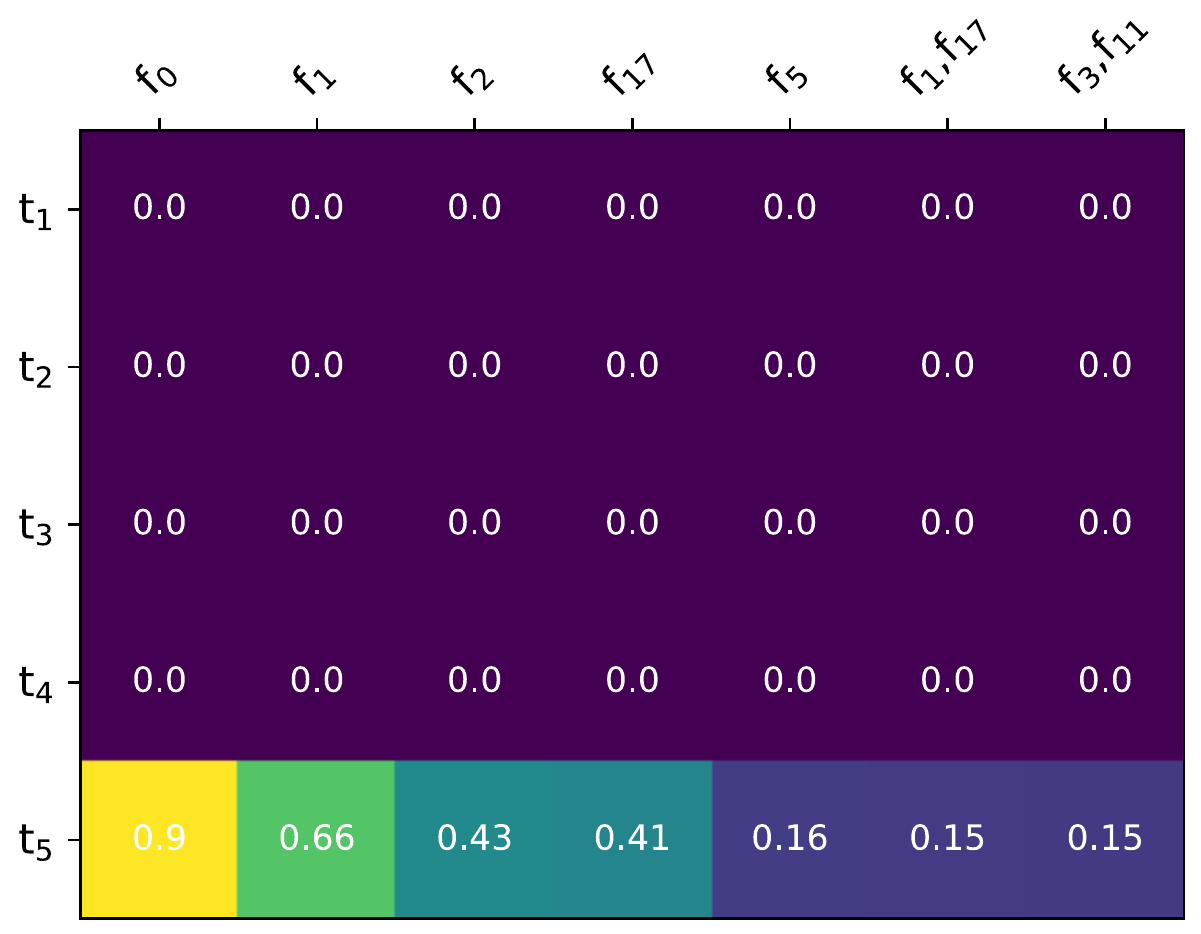}\vspace{1em}
		\includegraphics[height=0.7\linewidth,width=0.9\textwidth]{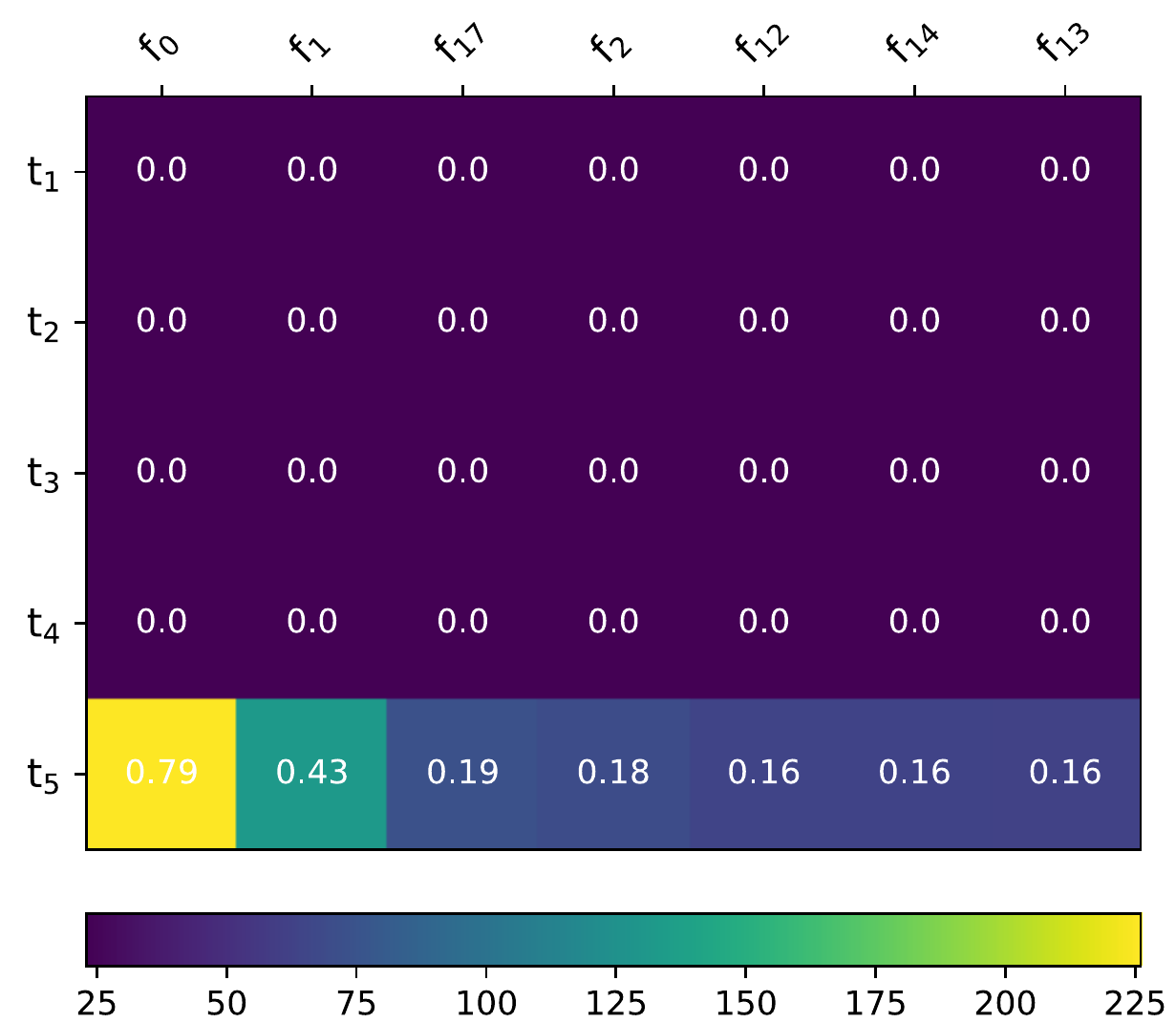}
	
		\caption*{\small (b) Negative instances. \vspace{-2mm}}
	\end{minipage}
	\vspace{-3mm}
	\caption{\small Heatmap examples of input feature weights w.r.t time points. The weights were first normalized to $[0,1]$, and then mapped to the color space $[0,255]$. The intensity of the color blocks indicates the importance of the corresponding features. We show the seven most important features (the sum of weights on different time points is greater than others). Here, $f_i$ indicates a feature, which special semantics can be queried in the Appendix \ref{appA}. $t_i$ is a time point and $t_5$ is the predictive time point.}\label{fig5}
\end{figure*}

\subsection{Explainable Enterprise Credit Rating}
A good enterprise credit rating system can provide accuracy evaluations and good explainability of the model and the outputs. Here, we describe how the proposed model is able to explain the output results and the modeling process. Benefiting from the dual attention modules and the feature crossing modules, the proposed model not only can provide credit rating prediction for a given company, but also can generate static and personalized explanations for the modeling process and the output prediction results, respectively.

Due to the PCA layers and the lasso regularizations in the feature crossing modules, we can summarize the most important feature combinations in different ranks of feature sets for a given dataset. As shown in Table~\ref{tab3}, we parsed the useful combination patterns and their weights from each PCA layer in a backtracking way (Section \ref{explanation}). By analyzing these static explanations, human experts in the field of enterprise credit rating can further investigate the trained model to determine whether there exists bias caused by the training datasets. For example, in the second-rank feature combinations, the combination of operation cycle (i.e., $f_{11}$) and capital return (i.e., $f_{18}$) may be an indication of a company's debt paying ability, and this pattern is consistent with common accounting principles. In addition, as shown in Table~\ref{tab3}, as the rank increases, the weights of the feature combinations tend to be trivial. This phenomenon proves that the high-rank features typically include a lot of redundant and noise, thus, the feature selection in the proposed feature crossing module is necessary.

In addition, in the field of enterprise credit rating task, users are interested in the correlations between the features of a given sample and the specific credit rating results. Therefore, we further provide a credit rating to a given company with visualized the correlations between the most important feature combinations and time points. We observe the following phenomenon from the visualized results shown in Fig.~\ref{fig5}: (1) The features of the time points closer to the rating time point are more important to the result than the features farther away from the rating time point, e.g., the features on $t_5$ are most effective to both positive samples and negative samples. (2) The results of different samples are affected by different features. (3) Some features are important to both positive samples and negative samples, e.g., the net profit cut growth rate (i.e., $f_{13}$) can discriminate both positive and negative samples.

In summary, the proposed model demonstrates good explainability and provide both global and personalized explanations, which can assist financial experts to assess the reliability of the trained models and users to understand the rating results.

\section{Conclusion}
In this paper, we have proposed a novel deep feature crossing based model to predict enterprise credit rations with high accuracy and explainability. The proposed model maps the original sparse and high-dimensional features into low-dimensional spaces and explicitly models the interactions of the high-rank feature. First, we construct and stack multiple feature crossing modules to generate useful high-rank feature combinations in an explicit manner. Then, by leveraging the proposed feature crossing modules, we learn static patterns of the high-rank feature combinations from the training data, which helps human experts in the field of credit rating identify bias in the trained model. Next, to obtain an accurate estimation with individual explanations, we further construct feature attention and temporal attention modules following the feature crossing stage. These attention modules are used to model the correlations of feature combinations and their temporal dependence, respectively. By mining and visualizing the Cartesian product of their attentions, the proposed method can provide personalized explanations of multiple feature combinations for a given sample and credit rating pair. In addition, we have presented methods to train the proposed model and generate corresponding explanations. The experimental results confirm that the proposed model demonstrates higher prediction accuracy than traditional enterprise credit rating models and provide explanations of both the prediction results and the training process.

However, the DNNs are a data-driven approach that can easily suffer over-fitting on small training sets. Therefore, in future, we plan to extend our experimental datasets to include non-listed companies and extend the proposed model to support non-financial information, which can effectively mitigate the problem of insufficient numbers of training samples. In addition, we would like to obtain the more accurate company representation by considering the news about companies and their propagation on social media, e.g., Weibo and Twitter to improve the performance of the proposed model.


%
\appendices
\renewcommand\thetable{\Alph{section}\arabic{table}}    
\section{}\label{appA}
The raw features of the listed Chinese companies used in our experiments are listed in Table \ref{tab4}.

\setcounter{table}{0}
\begin{table}[!htbp]\scriptsize
	\centering
	\caption {\small Attribute names of CH-Stocks dataset. $f_i$ indicates the feature number in our explanation system.\vspace{-2mm}}
	\begin{tabular}{|c|c|c|c|}
		\specialrule{0.2em}{1.1pt}{1.1pt}
		\multirow{2}{*}{NO.}&
		\multirow{2}{*}{Annotation}&
		\multirow{2}{*}{NO.}&
		\multirow{2}{*}{Annotation}\\ 
		\specialrule{0em}{1.1pt}{1.1pt}
		~&~&~&~\\\hline
		\specialrule{0em}{1.1pt}{1.1pt}
		$f_{0}$&Industry category&$f_{1}$&Net profit\\
		\specialrule{0em}{1.1pt}{1.1pt}
		$f_{2}$&Net profit cut&$f_{3}$&Gross revenue\\
		\specialrule{0em}{1.1pt}{1.1pt}
		$f_{4}$&Earnings per share&$f_{5}$&Net assets value per share\\
		\specialrule{0em}{1.1pt}{1.1pt}
		$f_{6}$&Capital surplus fund per share&$f_{7}$&Undivided profit per share\\
		\specialrule{0em}{1.1pt}{1.1pt}
		$f_{8}$&Operation cash flow per share&$f_{9}$&Days sales of inventory\\ 
		\specialrule{0em}{1.1pt}{1.1pt}
		$f_{10}$&Accounts receivable turnover days&$f_{11}$&Operation cycle\\
		\specialrule{0em}{1.1pt}{1.1pt}
		$f_{12}$&Net profit growth rate&$f_{13}$&Net profit cut growth rate\\
		\specialrule{0em}{1.1pt}{1.1pt}
		$f_{14}$&Operation revenue growth rate&$f_{15}$&Net profit ratio\\
		\specialrule{0em}{1.1pt}{1.1pt}
		$f_{17}$&Gross income ratio&$f_{18}$&Capital return\\
		\specialrule{0em}{1.1pt}{1.1pt}
		$f_{19}$&Return on equity&$f_{20}$&Inventory turning rate\\
		\specialrule{0em}{1.1pt}{1.1pt}
		$f_{21}$&Current ratio&$f_{22}$&Quick ratio\\
		\specialrule{0em}{1.1pt}{1.1pt}
		$f_{23}$&Super quick ratio&$f_{24}$&Debt equity ratio\\
		\specialrule{0em}{1.1pt}{1.1pt}
		$f_{25}$&Debt assets ratio&---&---\\
		\specialrule{0.1em}{1.1pt}{1.1pt}
	\end{tabular}
	\label{tab4}
\end{table}

\end{document}